\documentclass[lettersize,journal]{IEEEtran}
\usepackage{amsmath,amsfonts}
\usepackage{algorithmic}
\usepackage{algorithm}
\usepackage{array}
\usepackage[caption=false,font=normalsize,labelfont=sf,textfont=sf]{subfig}
\usepackage{textcomp}
\usepackage{stfloats}
\usepackage{url}
\usepackage{verbatim}
\usepackage{graphicx}
\usepackage{cite}

\usepackage{times}
\usepackage{epsfig}
\usepackage{amssymb}
\usepackage{multirow}
\usepackage{pifont}
\usepackage{comment}
\usepackage{xcolor}
\usepackage{hyperref}

\newcommand{\cmark}{\ding{51}}%
\newcommand{\xmark}{\ding{55}}%
\newcommand{\tabincell}[2]{\begin{tabular}{@{}#1@{}}#2\end{tabular}}

\def\etal{\emph{et al.~}}
\def\ie{\emph{i.e.}}
\def\eg{\emph{e.g.}} 
\def\etc{\emph{etc.}}

\begin{document}

\title{MonoIndoor++:Towards Better Practice of Self-Supervised Monocular Depth Estimation for Indoor Environments}

\author{Runze~Li,~\IEEEmembership{Student Member,~IEEE,}
        Pan~Ji$^\dagger$,
        Yi~Xu,
        ~Bir~Bhanu,~\IEEEmembership{Life~Fellow,~IEEE}
\thanks{Runze Li and Bir Bhanu are with the University of California Riverside, Riverside, CA, USA. Pan Ji and Yi Xu are with the OPPO US Research Center, InnoPeak Technology, Inc., USA. Part of this work was done while Runze Li was interning at OPPO US Research Center.}
\thanks{$^\dagger$Pan Ji (\href{mailto:peterji530@gmail.com}{peterji530@gmail.com}) is the corresponding author.}
}



\maketitle

\begin{abstract}
Self-supervised monocular depth estimation has seen significant progress in recent years, especially in outdoor environments, \ie, autonomous driving scenes. However, depth prediction results are not satisfying in indoor scenes where most of the existing data are captured with hand-held devices. As compared to outdoor environments, estimating depth of monocular videos for indoor environments, using self-supervised methods, results in two additional challenges: (i) the depth range of indoor video sequences varies a lot across different frames, making it difficult for the depth network to induce consistent depth cues for training, whereas the maximum distance in outdoor scenes mostly stays the same as the camera usually sees the sky; (ii) the indoor sequences recorded with handheld devices often contain much more rotational motions, which cause difficulties for the pose network to predict accurate relative camera poses, while the motions of outdoor sequences are pre-dominantly translational, especially for street-scene driving datasets such as KITTI. In this work, we propose a novel framework-\textbf{\textit{MonoIndoor++}} by giving special considerations to those challenges and consolidating a set of good practices for improving the performance of self-supervised monocular depth estimation for indoor environments. First, a depth factorization module with transformer-based scale regression network is proposed to estimate a global depth scale factor explicitly, and the predicted scale factor can indicate the maximum depth values. Second, rather than using a single-stage pose estimation strategy as in previous methods, we propose to utilize a residual pose estimation module to estimate relative camera poses across consecutive frames iteratively. Third, to incorporate extensive coordinates guidance for our residual pose estimation module, we propose to perform coordinate convolutional encoding directly over the inputs to pose networks. The proposed method is validated on a variety of benchmark indoor datasets, \ie, EuRoC MAV, NYUv2, ScanNet and 7-Scenes, demonstrating the state-of-the-art performance. In addition, the effectiveness of each module is shown through a carefully conducted ablation study and the good generalization and universality of our trained model is also demonstrated, specifically on ScanNet and 7-Scenes datasets.
\end{abstract}


\section{Introduction}
\label{sec:introduction}
\IEEEPARstart{M}{onocular} depth estimation has been applied in a variety of 3D perceptual tasks, including autonomous driving, virtual reality (VR), and augmented reality (AR). Estimating the depth map plays an essential role in these applications, in helping to understand environments, plan agents' motions, reconstruct 3D scenes, etc. Existing supervised depth methods~\cite{Eigen_2015_ICCV,Fu_2018_CVPR} can achieve high performance, but they require the ground-truth depth data during the training which is often expensive and time-consuming to obtain by using depth sensors (\eg, LiDAR). To this end, a number of recent work~\cite{garg2016unsupervised, zhou2017unsupervised, godard2019digging} have been focused on predicting the depth map from a single image using self-supervised manners and they have shown advantages in scenarios where obtaining the ground-truth is not possible. In these self-supervised frameworks, photometric consistency between multiple views from monocular video sequences has been utilized as the main supervision for training models. Specifically, the recent work~\cite{godard2019digging} 
has achieved significant success in estimating depth that is comparable to that by the supervised methods~\cite{guo2018learning,Fu_2018_CVPR}. For instance, on the KITTI dataset~\cite{Geiger2012CVPR}, the Monodepth2, proposed by Godard~\etal~\cite{godard2019digging}, achieves an absolute relative depth error (AbsRel) of 10.6\%, which is not far from the AbsRel of 7.2\% by the DORN which is a supervised model proposed by Fu~\etal~\cite{Fu_2018_CVPR}. However, most of these self-supervised depth prediction methods~\cite{garg2016unsupervised,zhou2017unsupervised,godard2019digging} are only evaluated on datasets of outdoor scenes such as KITTI, leaving their performance opaque for indoor environments. There are certainly ongoing efforts~\cite{zhou2019moving,zhao2020towards,bian2021tpami} which considers self-supervised monocular depth estimation for indoor environments, but their performance still trail far behind the one on the outdoor datasets by methods such as~\cite{garg2016unsupervised,zhou2017unsupervised,godard2019digging} or the supervised counterparts~\cite{Fu_2018_CVPR,Yin_2019_ICCV} on indoor datasets. 
In this paper, we concentrate on estimating the depth map from a single image for indoor environments in a self-supervised manner which only requires monocular video sequences for training. 

This paper investigates the performance discrepancies between the indoor and outdoor scenes and takes a step towards examining what makes indoor depth prediction more challenging than the outdoor case. We \textit{first} identify that the scene depth range of indoor video sequences varies a lot more than in the outdoor and conjecture that this posits more difficulties for the depth network in inducing consistent depth cues across images from monocular videos, resulting in the worse performance on indoor datasets. 

Our \textit{second} observation is that the pose network, which is commonly used in self-supervised methods~\cite{zhou2017unsupervised,godard2019digging}, tends to have large errors in predicting rotational parts of relative camera poses. A similar finding has been presented in ~\cite{zou2020learning} where predicted poses have much higher rotational errors (\eg, 10 times larger) than geometric SLAM~\cite{mur2017orb} even when they use a recurrent neural network as the backbone to model the long-term dependency for pose estimation. We argue that this problem is not prominent on outdoor datasets, \ie, KITTI, because the camera motions therein are mostly translational. However, frequent cameras rotations are inevitable in indoor monocular videos~\cite{Silberman:ECCV12,schonberger2016structure} as these datasets are often captured by hand-held cameras or Micro Aerial Vehicle (MAVs). Thus, the inaccurate rotation prediction becomes detrimental to the self-supervised training of a depth model for indoor environments. 

Our third conjecture is that the pose network in existing self-supervised methods is potentially suffering from insufficient cues to estimate relative cameras poses between color image pairs in different views. We argue that, rather than simply inducing camera poses based on color information of image pairs, encoding coordinates information can further improve the reliability of pose network in inferring geometric relations among changing views.

We propose {\bf MonoIndoor++}, a self-supervised monocular depth estimation method tailored for indoor environments, giving special considerations for above problems. Our MonoIndoor++ consists of three novel modules: a \textit{depth factorization} module, a \textit{residual pose estimation} module, and a \textit{coordinates convolutional encoding} module. In the depth factorization module, we factorize the depth map into a global depth scale (for the target image of the current view) and a relative depth map. The depth scale factor is separately predicted by an extra module (named as transformer-based scale regression network) in parallel with the depth network which predicts a relative depth map. In such a way, the depth network has more model plasticity to adapt to the depth scale changes during training. We leverage the recent advances of transformer~\cite{dosovitskiy2021an} in designing the scale regression network to predict the depth scale factor. In the residual pose estimation module, we mitigate the issue of inaccurate camera rotation prediction by performing residual pose estimation in addition to an initial large pose prediction. Such a residual approach leads to more accurate computation of the photometric loss~\cite{godard2019digging}, which in turn improves model training for the depth prediction. In the coordinates convolutional encoding module, we encode the coordinates information $(x,y)$ explicitly and incorporate them with color information in the residual pose estimation module, expecting to provide additional cues for pose predictions, which further consolidates residual pose estimation model during training.

It should be mentioned that this paper is an extended version of our previous conference paper~\cite{Ji_2021_ICCV}, where we propose an unsupervised learning framework for monocular depth estimation in indoor environments. In this paper, we \textbf{i)} add more technical details of our proposed method; \textbf{ii)} present \textit{coordinate convolutional encoding module} in the framework for improved performance of monocular depth prediction; \textbf{iii)} make a more clear explanation of our proposed 
\textit{depth factorization module} with \textit{transformer-based scale network}; \textbf{iv)} conduct extensive experiments and ablation studies on public benchmark datasets, \ie, EuRoC MAV~\cite{Burri25012016}, NYUv2~\cite{Silberman:ECCV12}, ScanNet~\cite{dai2017scannet} and 7-Scenes~\cite{Shotton_2013_CVPR}, and perform detailed analysis to demonstrate the effectiveness and good generalizability of our proposed framework in this journal paper.

In summary, our contributions are:
\begin{itemize}
    \item We propose a novel depth factorization module with a transformer-based scale regression network to estimate a global depth scale factor, which helps the depth network adapt to the rapid scale changes for indoor environments during model training.
    \item We propose a novel residual pose estimation module that mitigates the inaccurate camera rotation prediction issue in the pose network and in turn significantly improves monocular depth estimation performance.
    \item We incorporate coordinates convolutional encoding in the proposed residual pose estimation module to leverage coordinates cues in inducing relative camera poses.
    \item We demonstrate the state-of-the-art performance of self-supervised monocular depth prediction on a wide-variety of publicly available indoor datasets, \ie, EuRoC MAV~\cite{schonberger2016structure}, NYUv2~\cite{Silberman:ECCV12}, ScanNet~\cite{dai2017scannet} and 7-Scenes~\cite{Shotton_2013_CVPR}.
\end{itemize}

The paper is organized as follows: Section~\ref{sec:related_work} summarizes related published works in the field; then in Section~\ref{sec:method}, we explain our proposed approach for monocular depth estimation in indoor environments, the proposed approach consisting of a depth factorization, a residual pose and a coordinates convolutional encoding modules; and in Section~\ref{sec:experiments}, we present experimental results and ablation studies on a variety of benchmark indoor datasets; and lastly a conclusion of our work is discussed in Section~\ref{sec:conclusions}.

\section{Related Work}
\label{sec:related_work}
Much effort has been expended for the depth estimation in various environments. This paper addresses the self-supervised monocular depth estimation for indoor environments. In this section, we discuss the relevant work of depth estimation using both supervised and self-supervised methods.

\subsection{Supervised Monocular Depth Estimation}
The depth estimation problem was mostly solved by using supervised methods in early research. Saxena~\etal~\cite{saxena2008make3d} propose the method to regress the depth from a single image by extracting superpixel features and using a Markov Random Field (MRF). Schonberger~\etal~\cite{schnberger2016pixelwise} present a system for the joint estimation of depth and normal information with photometric and geometric priors. These methods employ traditional geometry-based methods. Eigen~\etal~\cite{eigen2014depth} propose the first deep-learning based method for monocular depth estimation using a multi-scale convolutional neural network (CNN). Later on, deep-learning based methods have shown significant progress on monocular depth estimation, specifically with massive ground-truth data during training the networks. One line of following work improves the performance of depth prediction by better network architecture design. Laina~\etal~\cite{laina2016deeper} propose an end-to-end fully convolutional architecture by encompassing the residual
learning to predict accurate single-view depth maps given monocular images. Bhat~\etal~\cite{Bhat_2021_CVPR} propose a transformer-based architecture block to adaptively estimate depth maps using a number of bins. Another line of work achieves improved depth estimation results by integrating more sophisticated training losses~\cite{Li_2017_ICCV,Fu_2018_CVPR,Yin_2019_ICCV, 9316778, 8764412, 8010878}. Besides, a few methods~\cite{ummenhofer2017demon,teed2018deepv2d} propose to use two networks, one for depth prediction and the other for motion, to mimic geometric Structure-from-Motion (SfM) or Simultaneous Localization and Mapping (SLAM) in a supervised framework. However, ground-truth depth maps with images are used to train these methods and obtaining ground-truth data is often expensive and time-consuming to capture. Some other methods then resort to remedy this problem by generating pseudo ground-truth depth labels with traditional 3D reconstruction methods~\cite{li2018megadepth,li2019learning}, such as SfM~\cite{schonberger2016structure} and SLAM~\cite{mur2017orb, teed2021droid}, or 3D movies~\cite{ranftl2019towards}. Such methods have better capacity of generalization across different datasets, but cannot necessarily achieve the best performance for the dataset at hand. Some other ongoing efforts explore to improve robustness of supervised monocular depth estimation for zero-shot cross-dataset transfer. Ranftl~\etal~\cite{ranftl2019towards} propose robust scale-and shift-invariant losses for training the model using data from mixed dataset and testing on zero-shot datasets, and improve it further by integrating vision transformer in network design~\cite{Ranftl_2021_ICCV}.


\subsection{Self-Supervised Monocular Depth Estimation}
Recently, significant progress has been made in self-supervised depth estimation as it does not require training with the ground-truth data. Garg~\etal~\cite{garg2016unsupervised} propose the first self-supervised method to train a CNN-based model for monocular depth estimation by using color consistency loss between stereo images. Zhou~\etal~\cite{zhou2017unsupervised} employ 
a depth network for depth estimation and a pose network to estimate relative camera poses between temporal frames, and use outputs to construct the photometric loss across temporal frames to train the model. Many follow-up methods then try to propose new training loss terms to improve self-supervision for training models. Godard~\etal~\cite{godard2017unsupervised} incorporate a left-right depth consistency loss for the stereo training. Bian~\etal~\cite{bian2019unsupervised} put forth a temporal depth consistency loss to ensure predicted depth maps of neighbouring frames are consistent. Wang~\etal~\cite{wang2018learning} first observe the diminishing issue of the depth model during training and propose a normalization method to counter this effect. Yin~\etal~\cite{yin2018geonet} and Zou~\etal~\cite{zou2018dfnet} train three networks (\ie, one depth network, one pose network, and one extra flow network) jointly by enforcing cross-task consistency between optical flow and dense depth. Wang~\etal~\cite{wang2019recurrent} and Zou~\etal~\cite{zou2020learning} explore techniques to improve the performance of pose network and/or the depth network by leveraging recurrent neural networks, such as LSTMs, to model long-term dependency. Tiwari~\etal~\cite{tiwari2020pseudo} design a self-improving loop with monocular SLAM and a self-supervised depth model~\cite{godard2019digging} to improve the performance of each one. 
Among these recent advances, Monodepth2~\cite{godard2019digging} significantly improves the performance over previous methods via a set of techniques: a per-pixel minimum photometric loss to handle occlusions, an auto-masking method to mask out static pixels, and a multi-scale depth estimation strategy to mitigate the texture-copying issue in depth. Watson~\etal~\cite{watson2021temporal} propose to use cost volume in the deep model and a new consistency loss calculated between the a teacher and a student model for self-supervision training. Unlike Monodepth2, this method show its advantages in using multiple frames during the testing. We implement our self-supervised depth estimation framework based on Monodepth2, but make important changes in designing both the depth and the pose networks.

Most of the aforementioned methods are only evaluated on outdoor datasets such as KITTI. Recent ongoing efforts~\cite{zhou2019moving,zhao2020towards,bian2020unsupervised} focus on self-supervised depth estimation for indoor environments. Zhou~\etal~\cite{zhou2019moving} first observe existing large rotations on most existing indoor datasets, and then use a pre-processing step to handle large rotational motions by removing all the image pairs with ``pure rotation'' and design an optical-flow based training paradigm using the processed data. Zhao~\etal~\cite{zhao2020towards} adopt a geometry-augmented strategy that solves for the depth via two-view triangulation and then uses the triangulated depth as supervision for model training. Bian~\etal~\cite{bian2020unsupervised, bian2021tpami} theoretically study the reasons behind the unsatisfying deep estimation performance in indoor environments and argue that ``the rotation behaves as noise during training''. They propose a rectification step during the data pre-precessing to remove the rotation between consecutive frames and design an auto-rectify network. We have an observation similar to~\cite{zhou2019moving, bian2020unsupervised} and~\cite{bian2021tpami} that large rotations cause difficulties for training the network. However, we take a different strategy. Instead of removing rotations from training data during the data pre-processing, we progressively estimate camera poses in rotations and translations via a novel residual pose module in an end-to-end manner, and we validate the effectiveness of the proposed method in predicting improved depth on a variety of indoor benchmark datasets.

\subsection{Transformer} We leverage the transformer in designing our scale regression network inspired by the recent advances~\cite{Wang_nonlocalCVPR2018,NIPS2017_3f5ee243,dosovitskiy2021an} of the attention mechanism. Self-attention in the transformer was first used successfully in natural lanuage processing (NLP) to model long-term dependencies. Wang~\etal~\cite{Wang_nonlocalCVPR2018} proposes a non-local operations for computer vision tasks. Recently, self-attention and its variants have been widely used in transformer networks for high-level visions tasks such
as image classification~\cite{dosovitskiy2021an} and semantic segmentation~\cite{liu2021Swin,liu2021swinv2}.

\subsection{Coordinates Encoding}
Convolutional neural networks (CNNs) have achieved significant success at many tasks, and it can be complemented with specialized layers for certain usage. For instance, detection models like Faster R-CNN~\cite{NIPS2015_14bfa6bb} make use of layers to compute coordinate transforms. Jaderberg~\etal~\cite{NIPS2015_33ceb07b} propose a spatial Transformer module that can be included into a standard CNN model to provide spatial transformation capabilities. Qi~\etal~\cite{qi2017pointnet,qi2017pointnetplusplus} design the PointNet which takes a set of 3D points represented as $(x,y,z)$ coordinates as well as extra color features for 3D classification and segmentation.
Recently, coordinates encoding has been widely used in vision transformers~\cite{dosovitskiy2021an,liu2021Swin,liu2021swinv2} and neural radiance fields representations~\cite{mildenhall2020nerf,park2021nerfies,tancik2022blocknerf}. Vision transformers~\cite{dosovitskiy2021an} take 2D images as the input, reshape the image into a sequence of flattened 2D patches and then employ self-attention blocks for image classification, detection and segmentation. Position embeddings are added to the patch embeddings as a standard processing step to retain positional information. Nerf~\cite{mildenhall2020nerf} proposes a method which takes a 3D location $(x,y,z)$ and 2D viewing direction $(\theta,\phi)$ as the input 
for scene synthesis. Unlike in vision transformers where positional encoding is utilized to provide discrete positions of tokens in the sequence, in Nerf, positional functions are used to map continuous input coordinates into a higher dimensional space for high frequency approximations. Liu~\etal~\cite{NEURIPS2018_60106888} define the \textit{CoordConv} operation to provide extra coordinates information as part of input channels to convolutional filters for the convolutional neural networks. Most of pose networks in monocular depth estimation pipelines~\cite{godard2017unsupervised,godard2019digging} simply take two consecutive frames as the input and outputs relative camera poses. We argue such designs infer rotational and translational relations by only focusing on photometric cues, but ignoring explicit coordinates cues. In our work, we leverage coordinates encoding in the proposed residual pose network.

\section{Method}
\label{sec:method}

\begin{figure*}[t]
\begin{center}
\includegraphics[width=1.0\textwidth]{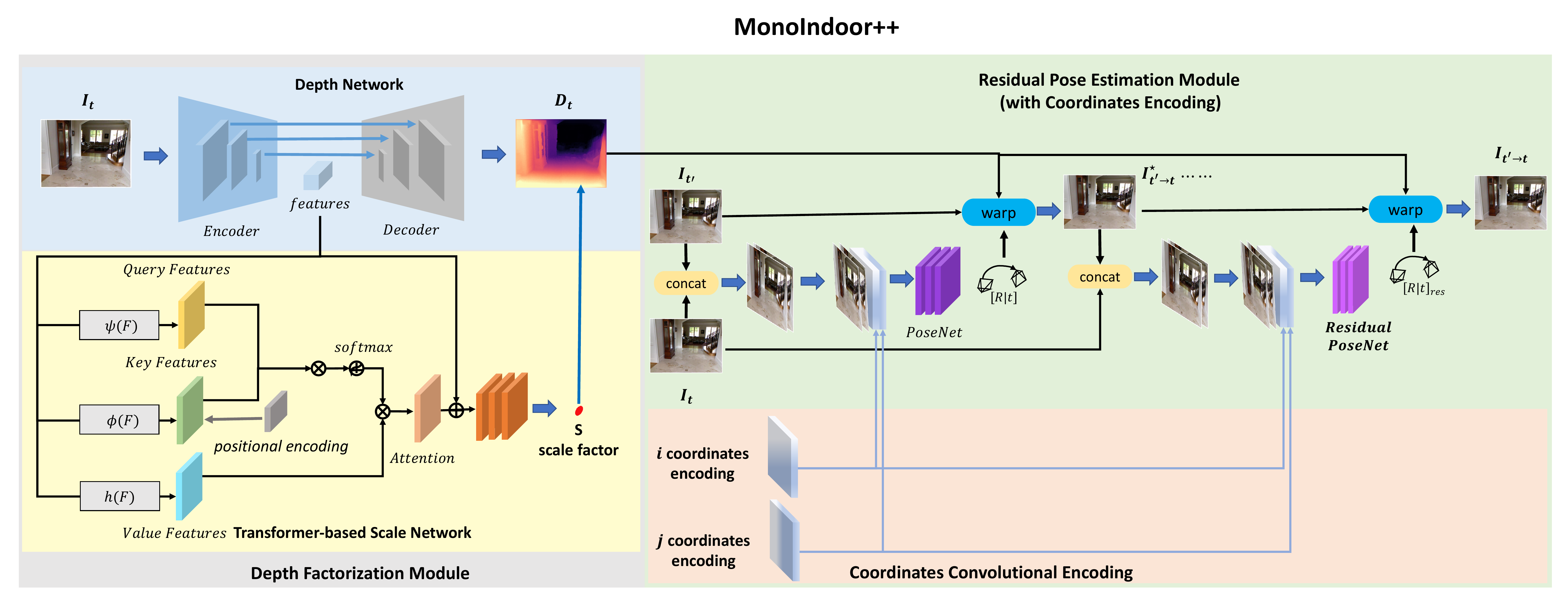}
\end{center}
\caption{Overview of the proposed \textbf{MonoIndoor++}. \textbf{Depth Factorization Module}: We use an encoder-decoder based depth network to predict a relative depth map and a transformer-based scale network to estimate a global scale factor. \textbf{Residual Pose Estimation Module}: We use a pose network to predict an initial camera pose of a pair of frames and residual pose network to iteratively predict residual camera poses based on the predicted initial pose. \textbf{Coordinates Convolutional Encoding}: We encode coordinates information along with the concatenated color image pairs as the input to the pose network and residual pose network for predicting relative camera poses.}
\label{fig:pipeline}
\end{figure*}

In this section, we present detailed descriptions of performing self-supervised depth estimation using the proposed \textbf{MonoIndoor++}. Specifically, we first give an overview of the standard framework for the self-supervised depth estimation. Then, we describe three core components including depth factorization, residual pose and coordinates convolution modules, respectively.  

\subsection{Self-Supervised Monocular Depth Estimation}
\label{sec:depth_estimation}
Self-supervised monocular depth estimation is considered as a novel view-synthesis problem which is defined in~\cite{zhou2017unsupervised,godard2019digging,zou2020learning}. This key idea is to train a model to predict the target image from different viewpoints of source images. The image synthesis is achieved by using the depth map as the bridging variable between the depth network and pose network. Both the depth map of the target image and the estimated relative camera pose between a pair of target and source images are required to train such systems. Specifically, the depth network predicts a dense depth map ${D}_t$ given a target image $I_t$ as the input. The pose network takes a target image $I_t$ and a source image $I_{t'}$ from another view and estimates a relative camera pose ${T}_{t \rightarrow t'}$ from the target to the source. The depth network and pose network are optimized jointly with the photometric reprojection loss which can then be constructed as follows:
\begin{equation}
    \mathcal{L}_{A} = \sum_{t'}\rho(I_t, I_{t'\rightarrow t}),
\end{equation}
and 
\begin{equation}
    I_{t'\rightarrow t} = I_{t'}\langle proj({D}_t, {T}_{t\rightarrow t'}, K) \rangle,
\label{eq:warp}
\end{equation}
where $\rho$ denotes the photometric reconstruction error~\cite{zhou2017unsupervised, godard2019digging}. It is a weighted combination of the L1 and Structured SIMilarity (SSIM) loss defined as 
\begin{equation}
    \rho(I_t, I_{t'\rightarrow t}) = \frac{\alpha}{2}\big(1-\texttt{\footnotesize SSIM}(I_t, I_{t'\rightarrow t})\big) + (1-\alpha)\|I_t, I_{t'\rightarrow t}\|_1.
    \label{eq:ssim_l1_loss}
\end{equation}
$I_{t'\rightarrow t}$ is the source image warped to the target coordinate frame based on the depth of the target image which is the output from the depth network. $proj()$ is the transformation function to map image coordinated $p_{t}$ from the target image to its $p_{t'}$ on the source image following
\begin{equation}
    p_{t'} \sim K{T}_{t\rightarrow t'}{D}_t(p_t)K^{-1}p_t,
    \label{eq:transform}
\end{equation}
and $\langle\cdot\rangle$ is the bilinear sampling operator which is locally sub-differentiable.

In addition, an edge-ware smoothness term is normally employed during training which can be written as 
\begin{equation}
    \mathcal{L}_{s} = |\partial_{x}d^{\ast}_{t}|e^{-|\partial_{x}I_t|}+|\partial_{y}d^{\ast}_{t}|e^{-|\partial_{y}I_t|},
\end{equation}
where $d^{\ast}_t = d/\bar{d}_t$ is the mean-normalized inverse depth from~\cite{wang2018learning}. 

Further, inspired by~\cite{bian2019unsupervised}, we incorporate the depth consistency loss to enforce the predicted depth maps across the target frame and neighbouring source frames to be consistent during the training. We first warp the predicted depth map ${D}_{t'}$ of the source image ${I}_{t'}$ by Equation~\eqref{eq:warp} to generate ${D}_{t'\rightarrow t}$, which is a corresponding depth map in the coordinate system of the source image. We then transform ${D}_{t'\rightarrow t}$ to the coordinate system of the target image via Equation~\eqref{eq:transform} to produce a synthesized target depth map $\widetilde{D}_{t'\rightarrow t}$. The depth consistency loss can be written as
\begin{equation}
    \mathcal{L}_d = \frac{|{D}_t - \widetilde{D}_{t'\rightarrow t}|} {{D}_t + \widetilde{D}_{t'\rightarrow t}}.
\end{equation}

Thus, the overall objective to train the model is
\begin{equation}
    \mathcal{L} = \mathcal{L}_{A} + \tau\mathcal{L}_{s} + \gamma \mathcal{L}_d,
\label{eq:photo_loss}
\end{equation}
where $\tau$ and $\gamma$ are the weights for the edge-aware smoothness loss and the depth consistency loss respectively.

As discussed in Section~\ref{sec:introduction}, existing self-supervised monocular depth estimation models have been used widely in producing competitive depth maps on datasets collected in outdoor environments, \eg, autonomous driving scenes. However, simply using these methods~\cite{godard2019digging} still suffer from worse performance in indoor environments, especially compared with fully-supervised depth prediction methods. We argue that the main challenges in indoor environments come from the fact that i) the depth range changes a lot and ii) indoor sequences captured in existing public indoor datasets, \eg, EuRoC MAV~\cite{Burri25012016} and NYUv2~\cite{Silberman:ECCV12}, contain regular rotational motions which are difficult to predict. To handle these issues, we propose \textbf{MonoIndoor++}, a self-supervised monocular depth estimation framework, as shown in Figure~\ref{fig:pipeline}, to enable improved predicted depth quality in indoor environments. The framework takes as input a single color image and outputs a depth map via our MonoIndoor++ which consists of two core parts: a depth factorization module with a transformer-based scale regression network and a residual pose estimation module. In addition, when designing the residual pose estimation, we incorporate coordinates convolutional operations to
encode coordinates information along with color information as input channels explicitly. The details of our main contributions are presented in the following sections.

\subsection{Depth Factorization Module}
\label{sec:depth_factorization}
Our depth factorization module consists of a depth prediction network and a transformer-based scale regression network. 
\subsubsection{Depth Prediction Network}
The backbone model of our depth prediction network is based on Monodepth2~\cite{godard2019digging}, which employs an auto-encoder structure with skip connections between the encoder and the decoder. The depth encoder learns a feature representation given a color image $I$ as input. The decoder takes features from the encoder as the input and outputs relative depth map prediction. In the decoder, a sigmoid activation function is used to process features from the last convolutional layers and a linear scaling function is utilized to obtain the final up-to-the-scale depth prediction, which can be written as follows,
\begin{equation}
d = 1/(a\sigma + b), 
\label{eq:depth_factor}
\end{equation}
where $\sigma$ is the outputs after the sigmoid function, $a$ and $b$ are specified to constrain the depth map ${D}$ within a certain depth range. $a$ and $b$ are pre-defined as a minimum depth value and a maximum depth value empirically according to a known environment. For instance, on the KITTI dataset ~\cite{Geiger2012CVPR} which is collected in outdoor scenes, $a$ is chosen as 0.1 and $b$ as 100. The reason for setting $a$ and $b$ as these fixed values is that the depth range is consistent across the video sequences when the camera always sees the sky at the far point. However, it is observed that this setting is not valid for most indoor environments. For instance, on the NYUv2 dataset~\cite{Silberman:ECCV12} which include various indoor scenes, \eg, office, kitchen, \etc, the depth range varies significantly as scene changes. Specifically, the depth range in a bathroom (\eg, 0.1m$\sim$3m) can be very different from the one in a lobby (\eg, 0.1m$\sim$10m). We argue that pre-setting depth range will act as an inaccurate guidance that is harmful for the model to capture accurate depth scales in training models. This is especially true when there are rapid scale changes, which are commonly observed on datasets~\cite{Silberman:ECCV12,Burri25012016,dai2017scannet} in indoor scenes. Therefore, to mitigate this problem, our depth factorization module learns a disentangled representation in the form of a relative depth map and a global scale factor. The relative depth map is obtained by the depth prediction network aforementioned and a global scale factor is outputted by a transformer-based scale regression network which is introduced in the next subsection.

\subsubsection{Transformer-based Scale Regression Network} We propose a transformer-based scale regression network (see Figure~\ref{fig:pipeline}) as a new branch which takes as input a color image and outputs its corresponding global scale factor. Our intuition is that the global scale factor can be informed by certain areas (\eg, the far point) in the images, and we propose to use a transformer block to learn the global scale factor. Our expectation is that the network can be guided to pay more attention to a certain area which is informative to induce the depth scale factor of the target image of the current view in a scene. 

The proposed transformer-based scale regression network takes the feature representations $\mathcal{F}\in\mathbb{R}^{D\times H\times W}$ learnt from the input image as the input and outputs the corresponding global scale factor, where $D$ is dimension, $H$ and $W$ are the height and width of the feature map. Specifically, we project input features $\mathcal{F}\in\mathbb{R}^{D\times H\times W}$ to the query, the key and the value output, which are defined as
\begin{equation}
  \begin{aligned}
  \psi(\mathcal{F}) &= \mathbf{W}_{\psi}\mathcal{F},
  \\
  \phi(\mathcal{F}) &= \mathbf{W}_{\phi}\mathcal{F},
  \\
  h(\mathcal{F}) &= \mathbf{W}_{h}\mathcal{F},
  \end{aligned}
\end{equation}
where $\mathbf{W}_{\psi}$, $\mathbf{W}_{\phi}$ and $\mathbf{W}_{h}$ are parameters to be learnt. The query and key values are then combined using the function $\mathcal{G_F}=\texttt{softmax}(\mathcal{F}^{T}\mathbf{W}^{T}_{\psi}\mathbf{W}_{\phi}\mathcal{F})h(\mathcal{F})$, giving the learnt self-attentions as $\mathcal{G_F}$. Finally, the $\mathcal{G_F}$ and the input feature representation $\mathcal{F}$ jointly contribute to the output $\mathcal{S_F}$ by using 
\begin{equation}
  \mathcal{S_F} = \mathbf{W}_{\mathcal{S_F}}\mathcal{G_F} + \mathcal{F}.
\end{equation}
Once we obtain $\mathcal{S_F}$, we apply two residual blocks including two convolutional layers in each, followed by three fully-connected layers with dropout layers in-between, to output the global scale factor $S$ for the target image of current view. We also use a 2D relative positional encoding~\cite{Bello_2019_ICCV} in calculating attentions with considerations of relative positional information of key features.

\subsubsection{Probabilistic Scale Regression Head}
The proposed transformer-based scale regression network is designed is to predict a single positive number given the input high-dimensional feature map $\mathcal{F}\in\mathbb{R}^{D\times H\times W}$. Inspired by the stereo matching work~\cite{chang2018pyramid}, we propose to use a probabilistic scale regression head to estimate the continuous value for scale factor. Specifically, given a maximum bound that the global scale factor is within, instead of outputting a single number directly, we first output a number of scale values $\widetilde{S}$ as the predictions of each scale $s$ and then calculate the probability of $s$ via the softmax operation $\texttt{softmax}(\cdot)$. Finally, the predicted global scale ${S}$ is calculated as the sum of each scale $s$ weighted by its probability of predicted values as
\begin{equation}
  {S} = \sum^{D_{max}}_{s=0}s\times \texttt{softmax}(\mathcal{\widetilde{S}}).
\end{equation}
Thus, the probabilistic scale regression head enables us to resolve regression problem smoothly with a probabilistic classification-based strategy (see Section~\ref{sec:ablation_depth_scale} for ablation results).


\subsection{Residual Pose Estimation}
\label{sec:residual_pose}
The principle of self-supervised monocular depth estimation is built upon the novel view synthesis, which requires both accurate depth maps from the depth network and camera poses from the pose network. Estimating accurate relative camera poses is important for calculating photometric reprojection loss to train the model because inaccurate camera poses might lead to wrong correspondences between the pixels in the target and source images, positing problems in predicting accurate depth maps. A standalone ``PoseNet" is widely used in existing methods~\cite{godard2019digging} to take two images as the input and to estimate the 6 Degrees-of-Freedom (DoF) relative camera poses. On datasets in outdoor environments (\eg, autonomous driving scenes like KITTI), we argue that the relative camera poses are fairly simple because the cars which are used to collect video data are mostly moving forward with large translations but minor rotations. This means that pose estimation is normally less challenging for the pose network. In contrast, in indoor environments, the video sequences in widely-used datasets~\cite{Silberman:ECCV12} are typically recorded with hand-held devices (\eg, Kinect), so there are more complicated ego-motions involved as well as much larger rotational motions. Thus, it is relatively more difficult for the pose network to learn to predict accurate relative camera poses. 

\begin{figure}[!t]
\begin{center}
\includegraphics[width=\linewidth]{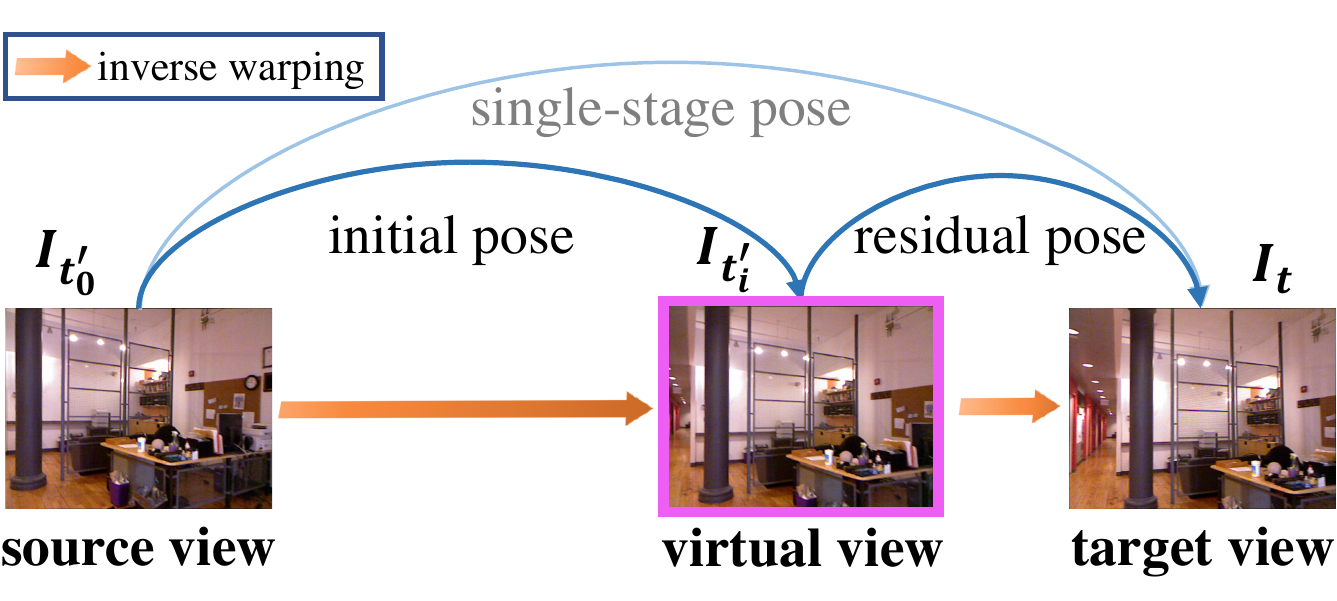}
\end{center}
\caption{Residual Pose Estimation. A single-stage pose can be decomposed into an {\it initial pose} and a {\it residual pose} by virtual view synthesis.}
\label{fig:residual_pose_pipeline}
\end{figure}

To better mitigate the aforementioned issues, existing methods ~\cite{zhou2019moving,bian2020unsupervised} concentrate on ``removing" or ``reducing" rotational components in camera poses during data preprocessing and train their models using preprocessed data. In this work, we argue these preprocessing techniques are not flexible in end-to-end training pipelines, instead, we propose a residual pose estimation module to learn the relative camera pose between the target and source images from different views in an iterative manner (see Figure~\ref{fig:residual_pose_pipeline} for core ideas). 

Our residual pose module consists of a standard pose network and a residual pose network. In the first stage, the pose network takes a target image $I_t$ and a source image $I_{t'_0}$ as input and predicts an initial camera pose ${T}_{t'_0\rightarrow t}$, where the subscript $0$ in $t'_{0}$ indicates that no transformation is applied over the source image yet. Then Equation~\eqref{eq:warp} is used to bilinearly sample from the source image, reconstructing a warped target image $I_{t'_{0}\rightarrow t}$ of a virtual view which is expected to be the same as the target image $I_t$ if the correspondences are solved accurately. However, it will not be the case due to inaccurate pose prediction. The transformation for this warping operation is defined as
\begin{equation}
    I_{t'_0\rightarrow t} = I_{t'}\langle proj({D}_t, {T}^{-1}_{t'_{0}\rightarrow t}, K) \rangle.
\label{eq:init_warp}
\end{equation}
Next, we propose a residual pose network (see \textbf{\textit{ResidualPoseNet}} in Figure~\ref{fig:pipeline}) which takes the target image and the synthesized target image of a virtual view ($I_{t'_{0}\rightarrow t}$) as input and outputs a residual camera pose ${T}^{res}_{(t'_{0}\rightarrow  t)\rightarrow t}$, representing the camera pose of the synthesized image $I_{t'_{0}\rightarrow t}$ with respect to the target image $I_{t}$.
Then, we bilinearly sample from the synthesized image as 
\begin{equation}
    I_{(t'_{0}\rightarrow t)\rightarrow t} = I_{t'_{0}\rightarrow t}\langle proj({D}_t, {T}^{res\;-1}_{(t'_0\rightarrow  t)\rightarrow t}, K) \rangle.
\label{eq:res_warp}
\end{equation}
Once a new synthesized image of a virtual view is obtained, we can continue to estimate the residual camera poses for next view synthesis operation.

We define the general form of Equation~\eqref{eq:res_warp} as
\begin{equation}
    I_{t'_{i}\rightarrow t} = I_{t'_{i}}\langle proj({D}_t, {T}^{res-1}_{t'_i\rightarrow t}, K) \rangle, i=0,1,\cdots\;.
\label{eq:res_warp_formal}
\end{equation}
by replacing the subscript $t'_{0}\rightarrow t$ with $t'_{1}$ to indicate that one warping transformation is applied, and similarly for the $i^{\rm th}$ transformation.

To this end, after multiple residual poses are estimated, the camera pose of source image $I_t'$ with respect to the target image $I_t$ can be written as ${T}_{t\rightarrow t'}={T}^{-1}_{t'\rightarrow t}$ where
\begin{equation}
    {T}_{t'\rightarrow t} = \prod_{i}{T}_{t'_i\rightarrow t}, i=\cdots,k,\cdots,1,0 \;.
\label{eq:pose}
\end{equation}

By iteratively estimating residual poses using a pose network and a residual pose network, we expect to obtain more accurate camera pose compared with the pose predicted from a single-stage pose network, so that a more accurate photometric reprojection loss can be built up for better depth prediction during the model training.

\subsection{Coordinates Convolutional Encoding}
\label{sec:coordinates_convolutional}
For self-supervised monocular depth estimation, most of existing methods are designed to induce relative camera poses given a pair of color images. In this work, we propose to incorporate coordinates information as a part of input channels along with the color information explicitly to provide additional coordinates cues for pose estimation. 

We extend standard convolutional layers to coordinates convolutional layers by initializing extra channels to process coordinates information which is concatenated channel-wise to the input representations (see \textbf{Coordinates Convolutional Encoding} in Figure~\ref{fig:pipeline}). Given a pair of 2D images, we encode two coordinates $x,y$ with color information $(r,g,b)$, resulting in the 8-channels input as $(r_1,g_1,b_1,r_2,g_2,b_2,i,j)$ where $(r_1,g_1,b_1)$ and $(r_2,g_2,b_2)$ are rgb values of color images, respectively. The $i$ coordinate channel is an $h\times w$ rank-1 matrix with its first row filled with 0's, its second row with 1's, its third with 2's, etc. The $j$ coordinate channel is similar, but with columns filled in with constant values instead of rows. A linear scaling operation is applied over both $i$ and $j$ coordinate values to encode them in the range $[-1, 1]$. We adopt coordinates convolutional layers~\cite{NEURIPS2018_60106888} in the residual pose estimation module to process 8-channels input for iterative pose estimation, and the pose estimation can be written as follows:
\begin{equation}
  T_{t\rightarrow t'}=RPModule(\mathbf{\Omega};Concat(I_{t},I_{t'},i,j))
\end{equation}
where $RPModule$ is the proposed pose estimation module, $\Omega$ is the parameters of the module which are to be optimized.

\section{Experiments}
\label{sec:experiments}

\subsection{Implementation Details}
We implement our model using PyTorch~\cite{NEURIPS2019_9015}. In the depth factorization module, we use the same depth network as in Monodepth2~\cite{godard2019digging}; for the transformer-based scale regression network, we use a transformer module followed by two basic residual blocks and then three fully-connected layers with a dropout layer in-between. The dropout rate is empirically set to 0.5. In the residual pose module, we let the residual pose networks use a common architecture as in Monodepth2~\cite{godard2019digging} which consists of a shared pose encoder and an independent pose regressor. In the coordinates encoding module, 2D coordinates information $(i,j)$ are directly concatenated with $(r,g,b)$ channels of color images as the input and the convolutional layers are initialized with ImageNet-pretrained weights. Each experiment is trained for 40 epochs using the Adam~\cite{kingma2015adam} optimizer and the learning rate is set to $10^{-4}$ for the first 20 epochs and it drops to $10^{-5}$ for remaining epochs. The smoothness term $\tau$ is set as 0.001. The consistency term $\gamma$ are set as 0.1 for EuRoC MAV dataset, 0.035 for NYUv2, ScanNet and 7-Scenes datasets, respectively.

\begin{table}[t!]
    \caption{Comparison of our method to existing supervised and self-supervised methods on NYUv2~\cite{Silberman:ECCV12}. Best results among supervised and self-supervised methods are in \textbf{bold}. }
    \label{tab:nyuv2_quan_full}
    \centering
    \resizebox{0.5\textwidth}{!}{
    \begin{tabular}{c|c|c|c|c|c|c}
    \hline
    \multirow{2}{*}{Methods} & 
    \multirow{2}{*}{Supervision} & \multicolumn{2}{c|}{Error Metric} & \multicolumn{3}{c}{Accuracy Metric} \\
    \cline{3-7}
    ~ & ~ & AbsRel & RMS & $\delta_1$ & $\delta_2$ & $\delta_3$ \\
    \hline
    Make3D~\cite{saxena2008make3d} & \cmark & 0.349 & 1.214 & 0.447 & 0.745 & 0.897 \\
    Depth Transfer~\cite{Karsch:TPAMI:14} & \cmark & 0.349 & 1.210 & - & - & - \\
    Liu~\etal~\cite{Liu_2014_CVPR} & \cmark & 0.335 & 1.060 & - & - & - \\
    Ladicky~\etal~\cite{Ladicky_2014_CVPR} & \cmark & - & - & 0.542 & 0.829 & 0.941 \\
    Li~\etal~\cite{Li_2015_CVPR} & \cmark & 0.232 & 0.821 & 0.621 & 0.886 & 0.968 \\
    Roy~\etal~\cite{Roy_2016_CVPR} & \cmark & 0.187 & 0.744 & - & - \\
    Liu~\etal~\cite{Liu_2015_CVPR} & \cmark & 0.213 & 0.759 & 0.650 & 0.906 & 0.976 \\
    Wang~\etal~\cite{Wang_2015_CVPR} & \cmark & 0.220 & 0.745 & 0.605 & 0.890 & 0.970 \\
    Eigen~\etal~\cite{Eigen_2015_ICCV} & \cmark & 0.158 & 0.641 & 0.769 & 0.950 & 0.988 \\
    Chakrabarti~\etal~\cite{NIPS2016_f3bd5ad5} & \cmark & 0.149 & 0.620 & 0.806 & 0.958 & 0.987 \\
    Laina~\etal~\cite{laina2016deeper} & \cmark & 0.127 & 0.573 & 0.811 & 0.953 & 0.988 \\
    Li~\etal~\cite{Li_2017_ICCV} & \cmark & 0.143 & 0.635 & 0.788 & 0.958 & 0.991 \\
    DORN~\cite{Fu_2018_CVPR} & \cmark & 0.115 & 0.509 & 0.828 & 0.965 & 0.992 \\
    Ranftl~\etal~\cite{Ranftl_2021_ICCV} & \cmark & 0.110 & \textbf{0.357} & 0.904 & \textbf{0.988} & 0.994 \\
    VNL~\cite{Yin_2019_ICCV} & \cmark & 0.108 & 0.416 & 0.875 & 0.976 & 0.994 \\
    Bhat~\etal~\cite{Bhat_2021_CVPR} & \cmark & 0.103 & 0.364 & \textbf{0.903} & 0.984 & \textbf{0.997} \\
    Fang~\etal~\cite{Fang_2020_WACV} & \cmark & \textbf{0.101} & 0.412 & 0.868 & 0.958 & 0.986 \\
    \hline
    \hline
    Zhou~\etal~\cite{zhou2019moving} & \xmark & 0.208 & 0.712 & 0.674 & 0.900 & 0.968 \\
    Zhao~\etal~\cite{zhao2020towards} & \xmark & 0.189 & 0.686 & 0.701 & 0.912 & 0.978 \\
    \tabincell{c}{Monodepth2~\cite{godard2019digging}} & \xmark & 0.160 & 0.601 & 0.767 & 0.949 & 0.988 \\
    \tabincell{c}{SC-Depth~\cite{bian2019unsupervised}} & \xmark & 0.159 & 0.608 & 0.772 & 0.939 & 0.982 \\
    P$^{2}$Net (3-frame)~\cite{IndoorSfMLearner} & \xmark & 0.159 & 0.599 & 0.772 & 0.942 & 0.984 \\
    P$^{2}$Net (5-frame)~\cite{IndoorSfMLearner} & \xmark & 0.147 & 0.553 & 0.801 & 0.951 & 0.987 \\
    Bian~\etal~\cite{bian2020unsupervised} & \xmark & 0.147 & 0.536 & 0.804 & 0.950 & 0.986 \\
    Bian~\etal~\cite{bian2021tpami} & \xmark & 0.138 & 0.532 & 0.820 & 0.956 & 0.989
    \\
    \hline
    \tabincell{c}{\textbf{MonoIndoor} \\ \textbf{(Ours)}} & \xmark & \textbf{0.132} & \textbf{0.517} & \textbf{0.834} & \textbf{0.961} & \textbf{0.990} \\
    \hline
    \end{tabular}
    }
\end{table}

\begin{figure*}[t!]
\begin{center}
\includegraphics[width=0.9\linewidth]{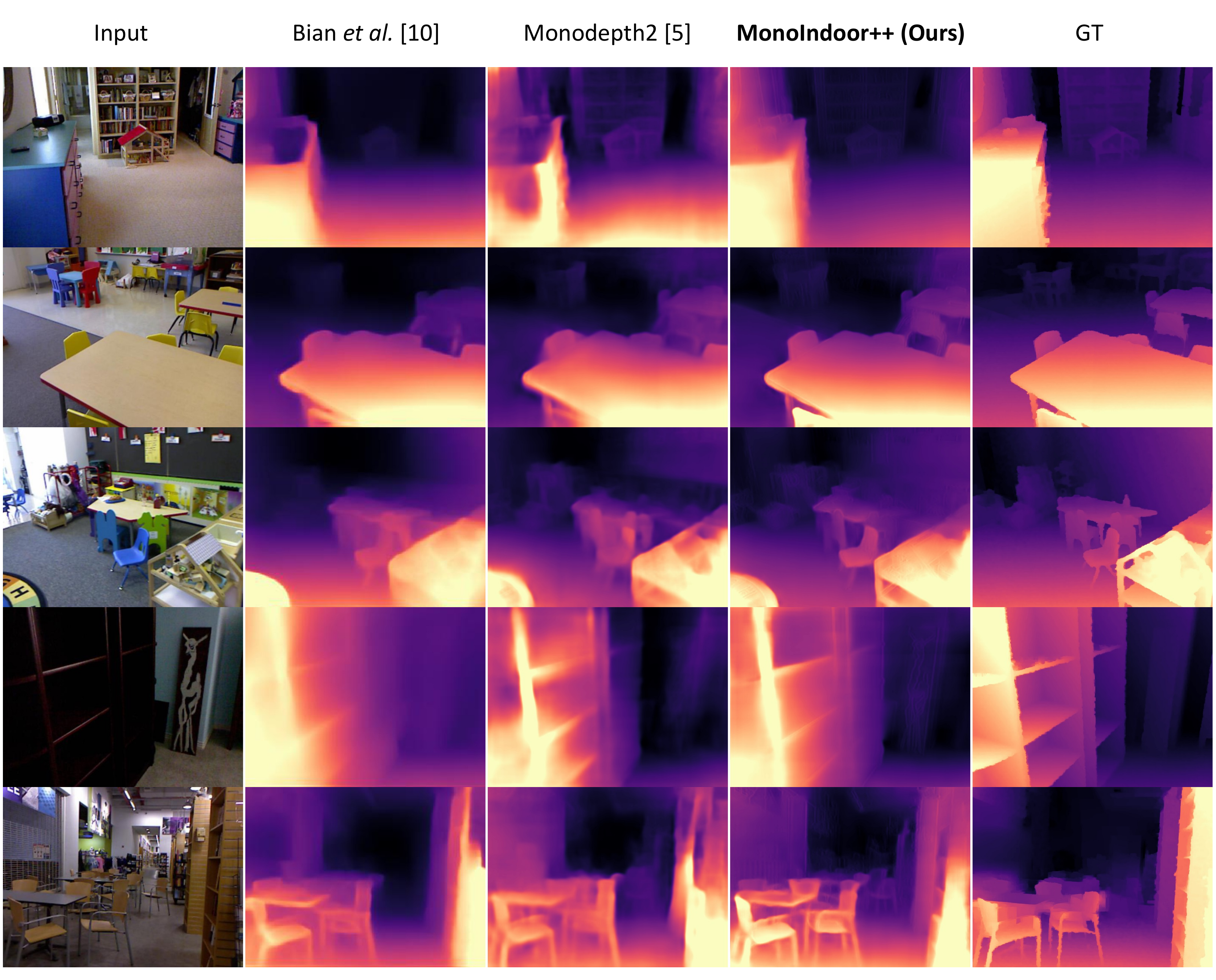}
\end{center}
\caption{Qualitative comparison on NYUv2~\cite{Silberman:ECCV12}. Images form the left to the right are: input, depth from ~\cite{bian2021tpami}, \cite{godard2019digging}, \textbf{MonoIndoor++(Ours)}, ground-truth depth. Compared with both the baseline method Monodepth2~\cite{godard2019digging} and recent work~\cite{bian2021tpami}, our model produces accurate depth maps that are closer to the ground-truth.}
\label{fig:nyuv2_qua_full}
\end{figure*}

\subsection{Datasets}
\subsubsection{NYUv2~\cite{Silberman:ECCV12}} The NYUv2 depth dataset contains 464 indoor video sequences which are captured by a hand-held Microsoft Kinect RGB-D camera. The dataset is widely used as a challenging benchmark for depth prediction. The resolution of videos is 640$\times$ 480. Images are rectified with provided camera intrinsics to remove image distortion. We use the official training and validation splits which include 302 and 33 sequences, respectively. We use officially provided 654 images with dense labelled depth maps for testing. During training, images are resized to 320$\times$256.

\subsubsection{EuRoC MAV~\cite{schonberger2016structure}}
The EuRoC MAV Dataset contains 11 video sequences captured in two main scenes, a machine hall and a vicon room. Sequences are categorized as \textit{easy}, \textit{medium} and \textit{difficult} according to the varying illumination and camera motions.
For the training, we use three sequences of ``Machine hall" (MH\_01, MH\_02, MH\_04) and two sequences of ``Vicon room'' (V1\_01 and V1\_02). Images are rectified with provided camera intrinsics to remove image distortion. During training, images are resized to 512$\times$256. We use the Vicon room sequence V1\_03, V2\_01, V2\_02 and V2\_03 for testing where the ground-truth depths are generated by projecting Vicon 3D scans onto the image planes. During training, images are resized to 512$\times$256. In addition, we use V2\_01 for ablation studies (see Section~\ref{sec:ablation_each_module} and Section~\ref{sec:ablation_depth_scale}).

\subsubsection{ScanNet~\cite{dai2017scannet}}
The ScanNet dataset contains RGB-D videos of 1513 indoor scans, which is captured by handheld devices. The dataset is annotated with 3D camera poses and instance-level semantic segmentations and is widely used on several 3D scene understanding tasks, including 3D object classification, semantic voxel labeling, and CAD model retrieval. We use officially released train-validation-test splits. The resolution of color images is $1296\times968$. During training, images are resized to 320$\times$256.

\subsubsection{7-Scenes~\cite{Shotton_2013_CVPR}} 7-Scenes dataset contains a number of video sequences captured in 7 different indoor scenes, \ie, \textit{office}, \textit{stairs}, etc. Each scene contains 500-1000 frames. All scenes are recorded using a handheld Kinect RGB-D camera at the resolution of $640\times480$. We use the official train-test split. During training, images are resized to 320$\times$256.

\subsection{Evaluation Metrics}
We use both error metrics and accuracy metrics proposed in~\cite{eigen2014depth} for evaluation on all datasets, which include the mean absolute relative error (AbsRel), root mean squared error (RMS) and the accuracy under threshold ($\delta_{i}<1.25^{i}, i=1,2,3$). Following previous self-supervised depth estimation methods~\cite{godard2019digging,zhao2020towards,bian2021tpami}, we multiply the predicted depth maps by a scalar that matches the median with that of the ground-truth because self-supervised monocular methods cannot recover the metric scale. The predicted depths are capped at 10m in all indoor datasets except the EuRoC MAV dataset which one is set as 20m because it contains ``Machine hall" scenes with observed large depth scale.

\subsection{Experimental Results}
\subsubsection{Results on NYUv2 Depth Dataset}
In this section, we evaluate our \textbf{MonoIndoor++} on the NYUv2 depth dataset~\cite{Silberman:ECCV12}. Following~\cite{zhao2020towards, bian2020unsupervised}, the raw dataset is firstly downsampled 10 times along the temporal dimension to remove redundant frames, resulting in $\sim20K$ images for training.

\noindent\textbf{Quantitative Results}
Table~\ref{tab:nyuv2_quan_full} presents the quantitative results of our model \textbf{MonoIndoor++} and both state-of-the-art (SOTA) supervised and self-supervised methods on NYUv2. It shows that our model outperforms previous self-supervised SOTA methods~\cite{godard2019digging,IndoorSfMLearner,bian2021tpami}, reaching the best results across all metrics. Specifically, compared with Monodepth2~\cite{godard2019digging}, which is our baseline method, our method improves monocular depth prediction performance significantly, reducing AbsRel from 16.0\% to \textbf{13.2\%} and increasing $\delta_1$ from 76.7\% to \textbf{83.4\%}. Besides, compared to a recent self-supervised method by Bian~\etal~\cite{bian2020unsupervised,bian2021tpami} which removes rotations via ``weak rectification'' with a data preprocessing step, our method gives the better performance without additional data preprocessing. It is noted that
NYUv2 is very challenging and many previous self-supervised
methods~\cite{yin2018geonet} fail to get satisfactory results. In addition to that, our model outperforms a group of supervised methods~\cite{Liu_2014_CVPR, Ladicky_2014_CVPR,Eigen_2015_ICCV,NIPS2016_f3bd5ad5,Li_2017_ICCV} and closes the performance gap between the self-supervised methods and fully-supervised methods~\cite{laina2016deeper,Fu_2018_CVPR}. 
\begin{table*}[!t]
    \caption{Quantitative results and comparisons between our \textbf{MonoIndoor++} by integrating our proposed core modules with the baseline Monodepth2~\cite{godard2019digging} on the test sequence V1\_03, V2\_01 V2\_02, V2\_03 of EuRoC. Best results are in \textbf{bold}.}
    \label{tab:eurco_quan_full}
    \centering
    \resizebox{1.0\textwidth}{!}{
    \begin{tabular}{c|c|c|c|c|c|c|c|c|c|c|c|c|c}
    \hline
    \multirow{3}{*}{Method}  & 
    \multirow{3}{*}{\tabincell{c}{Residual \\Pose}} & 
    \multirow{3}{*}{\tabincell{c}{Depth\\Factorization}} & \multirow{3}{*}{\tabincell{c}{Coordinates \\Encoding}} & \multicolumn{2}{c|}{Error Metric} &  \multicolumn{3}{c|}{Accuracy Metric} &
    \multicolumn{2}{c|}{Error Metric} &  \multicolumn{3}{c}{Accuracy Metric} \\
    \cline{5-14}
    ~ & ~ & ~ & ~ & AbsRel & RMSE  & $\delta_1$ & $\delta_2$ & $\delta_3$ & AbsRel & RMSE  & $\delta_1$ & $\delta_2$ & $\delta_3$ \\
    \cline{5-14}
     ~ & ~ & ~ & ~ & \multicolumn{5}{c|}{V1\_03} & \multicolumn{5}{c}{V2\_01} \\
    \hline
    Monodepth2~\cite{godard2019digging} & \xmark & \xmark & \xmark & 0.110 & 0.413 & 0.889 & 0.983 & 0.996 & 0.157 & 0.567 & 0.786 & 0.941 & 0.986 \\
    \textbf{MonoIndoor++ (Ours)} & \cmark & \xmark & \xmark & 0.100 & 0.379 & 0.913 & 0.987 & 0.997 & 0.141 & 0.518 & 0.815 & 0.961 & 0.991 \\
    \textbf{MonoIndoor++ (Ours)} & \cmark & \cmark & \xmark & 0.080 & 0.309 & 0.944 & 0.990 & 0.998 & 0.125 & 0.466 & 0.840 & 0.965 & 0.993 \\
    \textbf{MonoIndoor++ (Ours)} & \cmark & \cmark & \cmark & \textbf{0.079} & \textbf{0.303} & \textbf{0.949} & \textbf{0.991} & \textbf{0.998} & \textbf{0.115} & \textbf{0.439} & \textbf{0.861} & \textbf{0.972} & \textbf{0.992} \\
    \hline
    ~ & ~ & ~ & ~ & \multicolumn{5}{c|}{V2\_02} & \multicolumn{5}{c}{V2\_03} \\
    \hline
    Monodepth2~\cite{godard2019digging} & \xmark & \xmark & \xmark & 0.156  & 0.645  & 0.776 & 0.945 & 0.985 & 0.171 & 0.620 & 0.734 & 0.944 & 0.988\\
    \textbf{MonoIndoor++ (Ours)} & \cmark & \xmark & \xmark & 0.141 & 0.583 & 0.814 & 0.958 & 0.989 & 0.147 & 0.538 & 0.806 & 0.963 & 0.989 \\
    \textbf{MonoIndoor++ (Ours)} & \cmark & \cmark & \xmark & 0.142 & 0.581 & 0.802 & 0.952 & 0.990 & 0.140 & 0.502 & 0.810 & 0.964 & 0.993 \\
    \textbf{MonoIndoor++ (Ours)} & \cmark & \cmark & \cmark & \textbf{0.133} & \textbf{0.551} & \textbf{0.830} & \textbf{0.964} & \textbf{0.991} & \textbf{0.134} & \textbf{0.482} & \textbf{0.829} & \textbf{0.967} & \textbf{0.993} \\
    \hline
    \end{tabular}
    }
\end{table*}

\noindent\textbf{Qualitative Results} Figure~\ref{fig:nyuv2_qua_full} visualizes the predicted depth maps on NYUv2. Compared with the results from the baseline method Monodepth2~\cite{godard2019digging} and recent work~\cite{bian2021tpami}, depth maps predicted from our model (\textbf{MonoIndoor++}) are more precise and closer to the ground-truth. 
For instance, looking at the fourth column in the first row, the depth in the region of \textbf{\textit{cabinet}} predicted from our model is much sharper and cleaner, being close to the ground-truth (the last column). 
These qualitative results are consistent with our quantitative results in Table~\ref{tab:nyuv2_quan_full}.

\subsubsection{Results on EuRoC MAV Dataset}
In this section, we present evaluation of self-supervised monocular depth prediction and pose estimation. As there are not many public results on the EuRoC MAV dataset~\cite{schonberger2016structure}, we mainly compare between our \textbf{MonoIndoor++} and the baseline method Monodepth2~\cite{godard2019digging}.

\noindent\textbf{Quantitative Results}
We present and validate the effectiveness of each module of our \textbf{MonoIndoor++} and experimental results are shown Table~\ref{tab:eurco_quan_full}. 
It is observed that the depth prediction performance can be improved by adding each of our proposed modules. Our full model achieves the best performance across all evaluation metrics on various scenes including ``difficult" scenes ``Vicon room 103" and ``Vicon room 203". For instance, on ``Vicon room 201" (V2\_01), adding our residual pose module reduces the AbsRel from 15.7\% to 14.1\% and increases the $\delta_1$ from 78.6\% to 81.5\%, adding our depth factorization module reduces the AbsRel to 12.4\% and improves the $\delta_1$ to 84.0\% and adding our coordinates convolutional encoding reduces the AbsRel to \textbf{11.5\%} and increases the $\delta_1$ to \textbf{86.1\%} further. Similar improvements can be observed on other test sequences.

\begin{figure}[h]
\begin{center}
\includegraphics[width=\linewidth]{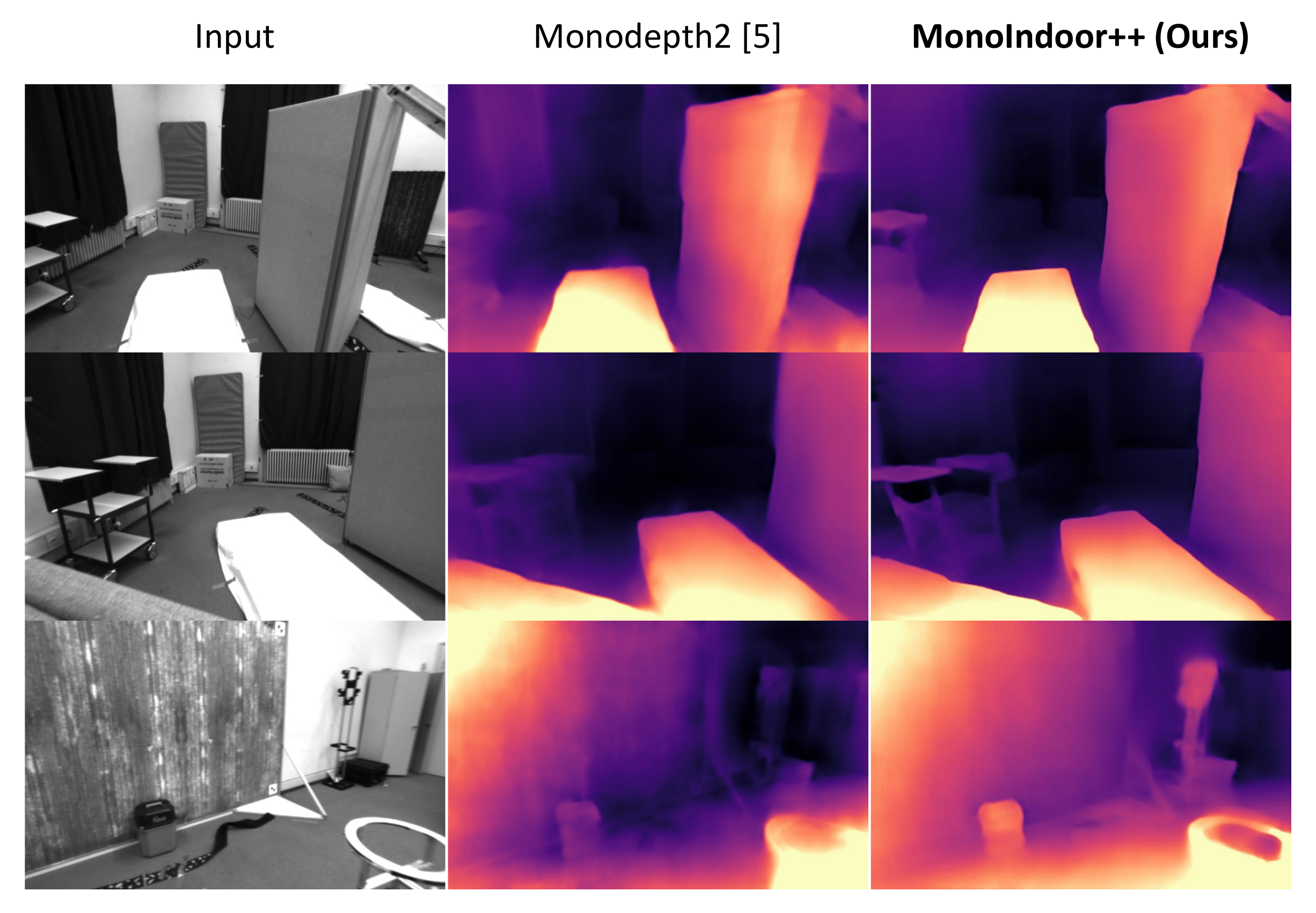}
\end{center}
\caption{Qualitative comparison of depth prediction on EuRoC MAV. Our \textbf{MonoIndoor++} produces more accurate and cleaner depth maps.}
\label{fig:euroc_qua_full}
\end{figure}

\noindent\textbf{Qualitative Results}
We present the qualitative results and comparisons of depth maps predicted by the baseline method Monodepth2~\cite{godard2019digging} and our \textbf{MonoIndoor++} in Figure~\ref{fig:euroc_qua_full}. There are no ground-truth dense depth maps on the EuRoC MAV dataset. From Figure~\ref{fig:euroc_qua_full}, it is clear that the depth maps generated by our model are much better than the ones by Monodepth2. For instance, in the third row, our model can predict precise depths for the \textbf{\textit{hole}} region at the right-bottom corner whereas such a hole structure in the depth map by Monodepth2 is missing. These observations are also consistent with the better quantitative results in Table~\ref{tab:eurco_quan_full}, proving the superiority of our model. 

\begin{table}[!h]
    \caption{Relative pose evalution on the EuRoC MAV~\cite{schonberger2016structure} dataset. Results show the average absolute trajectory error(ATE), and the relative pose error(RPE) in meters and degrees, respectively. Scene: test sequence name.}
    \label{tab:euroc_pose_eval}
    \centering
    \resizebox{\linewidth}{!}{
    \begin{tabular}{|c|c|c|c|c|}
    \hline
    Scene & Methods &
    ATE(m) & RPE(m) & RPE(\textdegree) \\
    \hline
    \multirow{2}{*}{V1\_03} & Monodepth2~\cite{godard2019digging} & 0.0681 & 0.0686 & 1.3237 \\
    \cline{2-5}
    ~ & \textbf{MonoIndoor++ (Ours)} & \textbf{0.0557} & \textbf{0.0542} & \textbf{0.5599} \\
    \hline
    \hline
    \multirow{2}{*}{V2\_01} & Monodepth2~\cite{godard2019digging} & 0.0266 & 0.0199 & 1.1985 \\
    \cline{2-5}
    ~ & \textbf{MonoIndoor++ (Ours)} & \textbf{0.0229} & \textbf{0.0050} & \textbf{1.1239} \\
    \hline
    \multirow{2}{*}{V2\_02} & Monodepth2~\cite{godard2019digging} & 0.0624 & 0.0481 & \textbf{6.4135} \\
    \cline{2-5}
    ~ & \textbf{MonoIndoor++ (Ours)} & \textbf{0.0517} & \textbf{0.0350} & 7.1928 \\
    \hline
    \multirow{2}{*}{V2\_03} & Monodepth2~\cite{godard2019digging} & 0.0670 & \textbf{0.0355} & 5.3532 \\
    \cline{2-5}
    ~ & \textbf{MonoIndoor++ (Ours)} & \textbf{0.0644} & 0.0676 & \textbf{4.8559} \\
    \hline
    \end{tabular}
    }
\end{table}

\begin{table*}[t!]
    \caption{Comparison of our method to latest self-supervised methods on RGB-D 7-Scenes~\cite{Shotton_2013_CVPR}. Best results are in \textbf{bold}}.
    \label{tab:7scenes_quan_full}
    \centering
    \resizebox{1.0\textwidth}{!}{
    \begin{tabular}{c|c|c|c|c|c|c|c|c|c|c|c|c}
    \hline
    \multirow{3}{*}{Scenes} &
    \multicolumn{4}{c|}{Bian~\etal~\cite{bian2020unsupervised}} &
    \multicolumn{4}{c|}{Monodepth2\cite{godard2019digging} (Baseline)} &
    \multicolumn{4}{c}{\textbf{MonoIndoor++ (Ours)}} \\
    \cline{2-13}
    ~ & \multicolumn{2}{c|}{Before Fine-tuning} &
    \multicolumn{2}{c|}{After Fine-tuning} &
    \multicolumn{2}{c|}{Before Fine-tuning} &
    \multicolumn{2}{c|}{After Fine-tuning} &
    \multicolumn{2}{c|}{Before Fine-tuning} &
    \multicolumn{2}{c}{After Fine-tuning} \\
    \cline{2-13}
    ~ & AbsRel & Acc $\delta_1$ & AbsRel & Acc $\delta_1$ & AbsRel & Acc $\delta_1$ & AbsRel & Acc $\delta_1$ & AbsRel & Acc $\delta_1$ & AbsRel & Acc $\delta_1$ \\
    \hline
    Chess & 0.169 & 0.719 & 0.103 & 0.880 &
    0.193 & 0.654 & 0.123 & 0.853 & 0.157 & 0.750 & \textbf{0.097} & \textbf{0.888} \\
    Fire & 0.158 & 0.758  & 0.089 & 0.916 &
    0.190 & 0.670 & 0.091 & 0.927 & 0.150 & 0.768 & \textbf{0.077} & \textbf{0.939} \\
    Heads & 0.162 & 0.749 & 0.124 & 0.862 &
    0.206 & 0.661 & 0.130 & 0.855 & 0.171 & 0.727 & \textbf{0.106} & \textbf{0.889} \\
    Office & 0.132 & 0.833 & 0.096 & 0.912 &
    0.168 & 0.748 & 0.105 & 0.897 & 0.130 & 0.837 & \textbf{0.083} & \textbf{0.934} \\
    Pumpkin & 0.117 & 0.857 & 0.083 & \textbf{0.946} &
    0.135 & 0.816 & 0.116 & 0.877 & 0.102 & 0.895 & \textbf{0.078} & 0.945 \\
    RedKitchen & 0.151 & 0.78 & 0.101 & 0.896 & 0.168 & 0.733 & 0.108 & 0.884 & 0.144 & 0.795 & \textbf{0.094} & \textbf{0.915} \\
    Stairs & 0.162 & 0.765 & 0.106 & 0.855 &
    0.146 & 0.806 & 0.127 & 0.825 & 0.155 & 0.753 & \textbf{0.104} & \textbf{0.857} \\
    \hline
    \end{tabular}
    }
\end{table*}

\noindent\textbf{Pose Evaluation}
In Table~\ref{tab:euroc_pose_eval}, we evaluate the proposed residual pose estimation module on all test sequences V1\_03, V2\_01, V2\_02 and V2\_03 of the EuRoC MAV~\cite{schonberger2016structure}. We follow~\cite{zhan2019dfvo} to evaluate relative camera poses estimated by our residual pose estimation module. We use the following evaluation metrics: absolute trajectory error (ATE) which measures the root-mean square error between predicted camera poses and ground-truth, and relative pose error (RPE) which measures frame-to-frame relative pose error in meters and degrees, respectively. As shown in Table~\ref{tab:euroc_pose_eval}, compared with the baseline model Monodepth2~\cite{godard2019digging} which employs one-stage pose network, using our method leads to improved relative pose estimation across evaluation metrics on most test scenes. Specifically, on the scene V1\_03, the ATE by our MonoIndoor is significantly decreased from
0.0681 meters to \textbf{0.0557} meters and RPE(\textdegree) is reduced from 1.3237\textdegree ~to \textbf{0.5599\textdegree}. 
Similar observations are made on the scene V2\_02, where the ATE by our MonoIndoor++ is significantly decreased from 0.0681 meters to \textbf{0.0557} meters.

\subsubsection{Results on RGB-D 7-Scenes Dataset}
In this section, we evaluate our \textbf{MonoIndoor++} on the RGB-D 7-Scenes dataset \cite{Shotton_2013_CVPR}. Following~\cite{bian2020unsupervised}, we extract one image from every 30 frames in each video sequence. For training, we first pre-train our model on the NYUv2 dataset, and then fine-tune the pre-trained model on each scene of 7-Scenes dataset.

Table~\ref{tab:7scenes_quan_full} presents the quantitative results and comparisons of our model \textbf{MonoIndoor++} and latest state-of-the-art (SOTA) self-supervised methods on 7-Scenes dataset. It can be observed that our model outperforms the baseline method Monodepth2~\cite{godard2019digging} significantly on each scene. Further, compared to the model~\cite{bian2020unsupervised}, our method achieve the best performance on most scenes before and after fine-tuning using NYUv2 pretrained models, which demonstrates better generalizability and capability of our model. Moreover, the results show that our method can perform well in a variety of different scenes.

\subsubsection{Results on ScanNet Dataset}
In this section, we evaluate our \textbf{MonoIndoor++} on the ScanNet dataset~\cite{dai2017scannet}. Referring to~\cite{tang2018ba}, the raw dataset is firstly downsampled 10 times along the temporal dimension and then $\sim100K$ images are randomly selected for training. During testing, $\sim4K$ are sampled from 100 different testing scenes to evaluate the trained model.

\noindent\textbf{Self-supervised Depth Estimation Evaluation}
Table~\ref{tab:scannet_quan_full} presents the quantitative results of our model \textbf{MonoIndoor++} and both state-of-the-art (SOTA) supervised and self-supervised methods on ScanNet. It shows that our model outperforms the previous self-supervised method~\cite{gu2021dro} in depth prediction, reaching the best results across all metrics. In addition to that, our model outperforms a group of supervised methods~\cite{ummenhofer_demon,tang2018ba}. 

\begin{table}[!t]
    \caption{Comparison of our method to existing supervised and self-supervised methods on ScanNet~\cite{dai2017scannet}. Best results among supervised and self-supervised methods are in \textbf{bold}.}
    \label{tab:scannet_quan_full}
    \centering
    \resizebox{0.5\textwidth}{!}{
    \begin{tabular}{c|c|c|c|c|c}
    \hline
    Methods & 
    Supervision & AbsRel & SqRel & RMS & RMS\textsubscript{log} \\
    \hline
    Photometric BA~\cite{AlismailBL16b} & \cmark & 0.268 & 0.427 & 0.788 & 0.330 \\
    DeMoN~\cite{ummenhofer_demon} & \cmark & 0.231 & 0.520 & 0.761 & 0.289 \\
    BANet~\cite{tang2018ba} & \cmark & 0.161 & 0.092 & 0.346 & 0.214 \\
    DeepV2D~\cite{teed2018deepv2d} & \cmark & 0.069 & \textbf{0.018} & 0.196 & 0.099 \\
    NeuralRecon~\cite{sun2021neucon} & \cmark & \textbf{0.047} & 0.024 & \textbf{0.164} & \textbf{0.093} \\
    \hline
    \hline
    Gu~\etal~\cite{gu2021dro} & \xmark & 0.140 & 0.127 & 0.496 & 0.212 \\
    \textbf{MonoIndoor++ (Ours)} & \xmark & \textbf{0.113} & \textbf{0.048} & \textbf{0.302} & \textbf{0.148} \\
    \hline
    \end{tabular}
    }
\end{table}

\noindent\textbf{Zero-shot Generalization}
We present the zero-shot generalization results of self-supervised depth prediction on ScanNet~\cite{dai2017scannet} in Table~\ref{tab:scannet_zero_shot_depth}, where we evaluate the proposed \textbf{MonoIndoor++} pretrained on NYUv2 dataset. From Table~\ref{tab:scannet_zero_shot_depth}, it is observed that our NYUv2 pretrained model generalizes better than other recent methods to new dataset. Besides, we show the zero-shot generalization results of relative pose estimation on ScanNet in Table~\ref{tab:scannet_zero_shot_pose}. We follow~\cite{IndoorSfMLearner,teed2018deepv2d} to use 2000 image pairs selected from diverse indoor scenes for pose evaluation. It can be observed that our method outperforms other self-supervised methods. Specifically, compared to Bian~\etal~\cite{bian2021tpami}, our method significantly reduces tr (cm) from 0.55 meters to \textbf{0.27} meters and decreases rot (deg) from 1.82 to \textbf{1.19}. Both depth and pose results validate the good zero-shot generalizability and capability of our method.

\begin{table}[!t]
    \caption{Zero-shot generalization of our method for self-supervised depth estimation on ScanNet~\cite{dai2017scannet}. Best results are in \textbf{bold}.}
    \label{tab:scannet_zero_shot_depth}
    \centering
    \resizebox{0.5\textwidth}{!}{
    \begin{tabular}{c|c|c|c|c|c|c}
    \hline
    \multirow{2}{*}{Methods} & 
    \multirow{2}{*}{Supervision} & \multicolumn{2}{c|}{Error Metric} & \multicolumn{3}{c}{Accuracy Metric} \\
    \cline{3-7}
    ~ & ~ & AbsRel & RMS & $\delta_1$ & $\delta_2$ & $\delta_3$ \\
    \hline
    Latina~\etal~\cite{laina2016deeper} & \cmark & 0.141 & 0.339 & 0.811 & .958 & 0.990 \\
    VNL~\cite{Yin_2019_ICCV} & \cmark & \textbf{0.123} & \textbf{0.306} & \textbf{0.848} & \textbf{0.964} & \textbf{0.991} \\
    \hline
    Zhao~\etal~\cite{zhao2020towards} & \xmark & 0.179 & 0.415 & 0.726 & 0.927 & 0.980 \\
    Bian~\etal~\cite{bian2019unsupervised} & \xmark & 0.169 & 0.392 & 0.749 & 0.938 & 0.983 \\
    Bian~\etal~\cite{bian2021tpami} & \xmark & 0.156 & 0.361 & 0.781 & 0.947 & 0.987 \\
    \hline
    Monodepth2~\cite{godard2019digging} & \xmark & 0.170 & 0.401 & 0.730 & 0.948 & 0.991 \\
    \textbf{MonoIndoor++ (Ours)} & \xmark & \textbf{0.138} & \textbf{0.347} & \textbf{0.810} & \textbf{0.967} & \textbf{0.993} \\
    \hline
    \end{tabular}
    }
\end{table}

\begin{table}[!t]
    \caption{Zero-shot generalization of our method for relative pose estimation on ScanNet~\cite{dai2017scannet}. Best results are in \textbf{bold}.}
    \label{tab:scannet_zero_shot_pose}
    \centering
    \resizebox{0.5\textwidth}{!}{
    \begin{tabular}{c|c|c|c}
    \hline
    Methods & rot (deg) & tr (deg) & tr (cm) \\
    \hline
    Zhou~\etal~\cite{zhou2019moving}  & 1.96 & 39.17 & 1.4 \\
    Monodepth2~\cite{godard2019digging} & 2.03 & 41.12 & 0.83 \\
    P$^{2}$Net~\cite{IndoorSfMLearner}  & 1.86 & 35.11 & 0.89 \\
    Bian~\etal~\cite{bian2021tpami}  & 1.82 & 39.41 & 0.55 \\
    \hline
    \textbf{MonoIndoor++ (Ours)} & \textbf{1.19} & \textbf{21.33} & \textbf{0.27} \\
    \hline
    \end{tabular}
    }
\end{table}

\subsection{Ablation Studies}
\label{sec:ablation}
\subsubsection{Effects of each proposed module} 
\label{sec:ablation_each_module}

\begin{table*}[!t]
    \caption{Ablation results of our \textbf{MonoIndoor++} and comparison with the baseline method on the test sequence V2\_01 of EuRoC and NYUv2 datasets. Best results are in \textbf{bold}.}
    \label{tab:nyuv2_euroc_quan_ablation}
    \centering
    \resizebox{1.0\textwidth}{!}{
    \begin{tabular}{c|c|c|c|c|c|c|c|c|c|c|c|c|c|c}
    \hline
    \multirow{3}{*}{Method}  & 
    \multirow{3}{*}{\tabincell{c}{Residual \\Pose}} & 
    \multirow{3}{*}{\tabincell{c}{Depth\\Factorization}} & \multirow{3}{*}{\tabincell{c}{Coordinates \\Encoding}} & \multirow{3}{*}{\tabincell{c}{Depth\\Consistency}} &
    \multicolumn{5}{c|}{EuRoC MAV V2\_01} &  \multicolumn{5}{c}{NYUv2} \\
    \cline{6-14}
    ~ & ~ & ~ & ~ & ~ &
    \multicolumn{2}{c|}{Error Metric} &  \multicolumn{3}{c|}{Accuracy Metric} & \multicolumn{2}{c|}{Error Metric} &  \multicolumn{3}{c}{Accuracy Metric} \\
    \cline{6-14}
    ~ & ~ & ~ & ~ & ~ & AbsRel & RMSE  & $\delta_1$ & $\delta_2$ & $\delta_3$ & AbsRel & RMSE  & $\delta_1$ & $\delta_2$ & $\delta_3$ \\
    \hline
    Monodepth2~\cite{godard2019digging} & \xmark & \xmark & \xmark & \xmark & 0.157 & 0.567 & 0.786 & 0.941 & 0.986 & 0.16 & 0.601 & 0.767 & 0.949 & 0.988 \\
    \textbf{MonoIndoor++ (Ours)} & \cmark & \xmark & \xmark & \xmark & 0.141 & 0.518 & 0.815 & 0.961 & 0.991 & 0.142 & 0.553 & 0.813 & 0.958 & 0.988 \\
    \textbf{MonoIndoor++ (Ours)} & \cmark & \xmark & \cmark & \xmark & 0.129 & 0.481 & 0.843 & 0.971 & 0.990 & 0.139 & 0.545 & 0.821 & 0.958 & 0.989 \\ 
    \hline
    \textbf{MonoIndoor++ (Ours)} & \xmark & \cmark & \xmark & \xmark & 0.149 & 0.535 & 0.805 & 0.955 & 0.987 & 0.152 & 0.576 & 0.792 & 0.951 & 0.987 \\
    \textbf{MonoIndoor++ (Ours)} & \cmark & \cmark & \xmark & \xmark & 0.133 & 0.501 & 0.833 & 0.961 & 0.987 & 0.139 & 0.549 & 0.818 & 0.958 & 0.988 \\
    \textbf{MonoIndoor++ (Ours)} & \cmark & \cmark & \xmark & \cmark & 0.125 & 0.466 & 0.840 & 0.965 & 0.993 & 0.134 & 0.526 & 0.823 & 0.958 & 0.989 \\
    \hline
    \textbf{MonoIndoor++ (Ours)} & \cmark & \cmark & \cmark & \cmark & \textbf{0.115} & \textbf{0.439} & \textbf{0.861} & \textbf{0.972} & \textbf{0.992} & \textbf{0.132} & \textbf{0.517} & \textbf{0.834} & \textbf{0.961} & \textbf{0.990} \\
    \hline
    \end{tabular}
    }
\end{table*}

\begin{figure*}[t]
\begin{center}
\includegraphics[width=\linewidth]{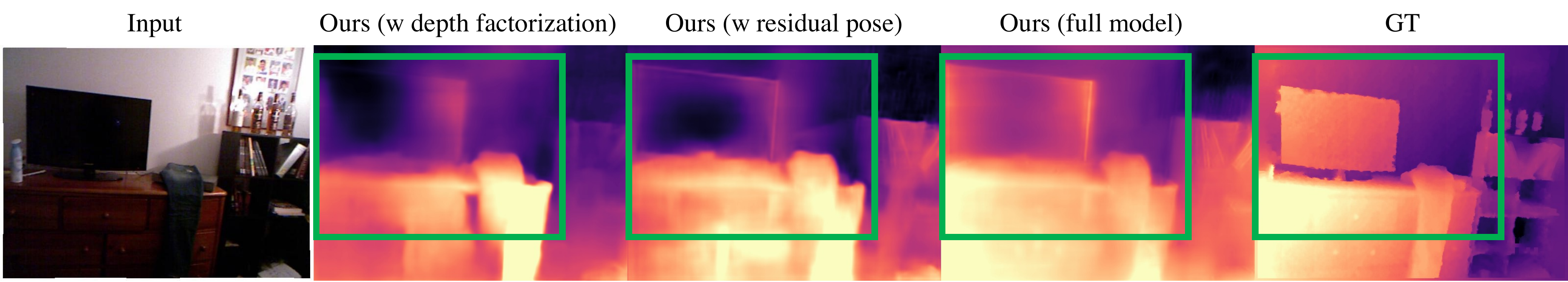}
\end{center}
\caption{Qualitative ablation comparisons of depth prediction on NYUv2. Our full model with both depth factorization and residual pose modules produce better depth maps.}
\label{fig:nyuv2_qua_ablations}
\end{figure*}

\begin{figure*}[t]
\begin{center}
\includegraphics[width=\linewidth]{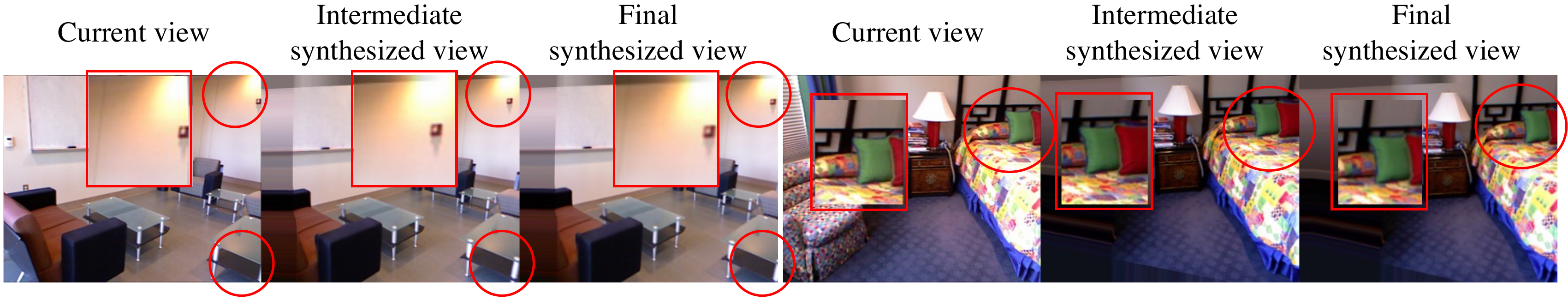}
\end{center}
\caption{Intermediate synthesized views on NYUv2.}
\label{fig:nyuv2_intermediate_views}
\end{figure*}

In this section, we perform ablation studies for each module in our proposed \textbf{MonoIndoor++} on the NYUv2 dataset~\cite{Silberman:ECCV12} and the test scene ``V2\_01" of the EuRoC MAV dataset~\cite{schonberger2016structure} in Table~\ref{tab:nyuv2_euroc_quan_ablation}. 

Specifically, We first perform ablation study for the residual pose estimation module. In Table~\ref{tab:nyuv2_euroc_quan_ablation} (see row 2 and 3), improved performance can be observed by using the residual pose module, decreasing the AbsRel from 15.7\% to 14.1\% and increasing $\delta_1$ to 81.5\% on EuRoC MAV V2\_01, and decreasing the AbsRel from 16\% to 14.2\% and increasing $\delta_1$ to 81.3\% on NYUv2.

Next, we experiment to validate the effectiveness of the depth factorization module (see row 2 and 5). Comparing with Monodepth2 which predicts depth without any guidance of global scales, by using the depth factorization module with a separate scale network, we observe an improved performance from 15.7\% down to 14.9\% for AbsRel and from 78.6\% up to 80.5\% for $\delta_1$ on EuRoC MAV V2\_01, and decreasing the AbsRel to 15.2\% and increasing $\delta_1$ to 79.2\% on NYUv2.

In addition, by applying the residual pose estimation module along with either the depth factorization module (row 6) or coordinates convolutional encoding (row 4), the performance can be improved consistently. Further, we present ablation results by using depth consistency loss with the residual pose estimation module and the depth factorization module (see row 6 and 7). Also, ablation results of our full model with and without coordinate convolutional encoding are shown in row 7 and row 8.

Last but not the least, using our full model \textbf{MonoIndoor++}, significant improvements can be achieved across all evaluation metrics. For instance, the AbsRel is reduced to \textbf{11.5\%} and the $\delta_1$ is increased to \textbf{86.1\%} on EuRoC MAV V2\_01 and the AbsRel is reduced to \textbf{13.2\%} and $\delta_1$ is increased to \textbf{83.2\%} on NYUv2. These ablation results clearly prove the effectiveness of the proposed modules. 

We also present the exemplar depth visualizations by our proposed modules on NYUv2 dataset in Figure~\ref{fig:nyuv2_qua_ablations}. In addition, we visualize intermediate and final synthesized views compared with the current view on NYUv2 in Figure~\ref{fig:nyuv2_intermediate_views}. Highlighted regions show that final synthesized views are better than the intermediate synthesized views and closer to the current view. 



\subsubsection{Effects of network design for transformer-based scale regression network}
\label{sec:ablation_depth_scale}
We perform ablation studies for our network design choices for the transformer-based scale regression network in depth factorization module on the test sequence V2\_01 of the EuRoC MAV dataset~\cite{Silberman:ECCV12}. Firstly, we consider the following designs as the backbone of our scale regression network: I) a pre-trained ResNet-18~\cite{He_2016_CVPR} followed by a group of Convolutional-BN-ReLU layers; II) a pre-trained ResNet-18~\cite{He_2016_CVPR} followed by two residual blocks; III) a lightweight network with two residual blocks which shares the feature maps from the depth encoder as input. These three choices are referred to as the ScaleCNN, ScaleNet and ScaleRegressor, respectively in Table~\ref{tab:eurco_quan_ablation}. Next, we validate the effectiveness of adding new components into our backbone design. As described in Section~\ref{sec:depth_factorization}, we mainly integrate two sub-modules: i) a transformer module and ii) a probabilistic scale regression block.

\begin{table}[!t]
    \caption{Ablation results of design choices and the effectiveness of components in the transformer-based scale regression network of our model (\textbf{MonoIndoor++}) on EuRoC MAV~\cite{schonberger2016structure}. Porb. Reg.: the probabilistic scale regression block. Note: we only use the residual pose estimation module when experimenting with different network designs for the depth factorization module.}
    \label{tab:eurco_quan_ablation}
    \centering
    \resizebox{0.5\textwidth}{!}{
    \begin{tabular}{c|c|c|c|c|c|c|c}
    \hline
    \multirow{2}{*}{Network Design} & 
    \multirow{2}{*}{Attention} &
    \multirow{2}{*}{\tabincell{c}{Prob.\\Reg.}} &
    \multicolumn{2}{c|}{Error Metric} &  \multicolumn{3}{c}{Accuracy Metric} \\
    \cline{4-8}
    ~ & ~ & ~ & AbsRel & RMSE  & $\delta_1$ & $\delta_2$ & $\delta_3$ \\
    \hline
    I. ScaleCNN & \cmark & \cmark & 0.140 & 0.518 & 0.821 & 0.956 & 0.985 \\
    II. ScaleNet & \cmark & \cmark & 0.141 & 0.519 & 0.817 & 0.959 & 0.988 \\
    \hline
    III. ScaleRegressor & \xmark & \xmark & 0.139 & 0.508 & 0.817 & 0.960 & 0.987 \\
    III. ScaleRegressor & \cmark & \xmark & 0.135 & 0.501 & 0.825 & 0.964 & 0.989 \\
    III. ScaleRegressor & \cmark & \cmark & \textbf{0.125} & \textbf{0.466} & \textbf{0.840} & \textbf{0.965} & \textbf{0.993} \\
    \hline
    \end{tabular}
    }
\end{table}

As shown in Table~\ref{tab:eurco_quan_ablation}, the best performance is achieved by ScaleRegressor that uses transformer module and probabilistic scale regression. It proves that sharing features with the depth encoder is beneficial to scale estimation. Comparing the results of three ScaleRegressor variants, the performance gradually improves as we add more components (\ie., attention and probabilistic scale regression (Prob. Reg.)). Specifically, adding the transformer module improves the overall performance over the baseline backbone; adding the probabilistic regression block leads to a further improvement, which validates the effectiveness of our proposed sub-modules.


\subsubsection{Ablation results of coordinates convolutional encoding}
\label{sec:ablation_coordinates_convolutional}
We present ablation studies for the encoding position of the coordinates convolutional encoding module on the NYUv2 dataset~\cite{Silberman:ECCV12} in Tabel~\ref{tab:nyuv2_coord_ablation}. To explore the effectiveness of using coordinates encoding technique, we only run our \textbf{MonoIndoor++} with the residual pose estimation module. We perform coordinates convolutional encoding with the following choices. Specifically, we first encode coordinates information with the color image pairs and extend coordinates convolutional layers to process combined input data. Second, we perform coordinates encoding operations with the feature representations outputted from the pose encoder and the processed features are taken as the input to the pose decoder. Third, we incorporate coordinates encoding operations with both input and features from pose encoder for pose estimation. 

\begin{table}[h!]
    \caption{Ablation results of encoding position for coordinates conovlutional with our \textbf{MonoIndoor++} on NYUv2. Init.: initialization of weights.}
    \label{tab:nyuv2_coord_ablation}
    \centering
    \resizebox{1.0\linewidth}{!}{
    \begin{tabular}{c|c|c|c|c|c|c}
    \hline
    \multirow{2}{*}{Model} &
    \multirow{2}{*}{\tabincell{c}{Encoding \\ Position}} & \multicolumn{2}{c|}{Error Metric} &  \multicolumn{3}{c}{Accuracy Metric} \\
    \cline{3-7}
    ~ & ~ & AbsRel & RMS & $\delta_1$ & $\delta_2$ & $\delta_3$ \\
    \hline
    \tabincell{c}{MonoIndoor } & \xmark & 0.142 & 0.553 & 0.813 & 0.958 & 0.988 \\
    \hline
    \tabincell{c}{\textbf{MonoIndoor++} \\ 
    (Random Init.)} & Input & 0.140 & \textbf{0.543} & 0.817 & 0.959 & 0.989 \\
    \hline
    \tabincell{c}{\textbf{MonoIndoor++} \\
    \textbf{(ImageNet Init.)}} & Input & \textbf{0.139} & 0.545 & \textbf{0.821} & \textbf{0.958} & \textbf{0.989} \\
    \hline
    \tabincell{c}{\textbf{MonoIndoor++} \\ (ImageNet Init.) } & Encoder Features & 0.145 & 0.565 & 0.806 & 0.954 & 0.988 \\
    \hline
    \tabincell{c}{\textbf{MonoIndoor++} \\ (ImageNet Init.)} & Input \& Encoder Features & 0.141 & 0.554 & 0.815 & 0.957 & 0.989 \\
    \hline
    \end{tabular}
    }
\end{table}

From the Table~\ref{tab:nyuv2_coord_ablation}, it can be observed that, by using coordinates convolutional encoding in residual pose estimation module, performance can be improved. For instance, the AbsRel is decreased to \textbf{13.9\%} from 14.2\% and the $\delta_1$ is improved from 81.7\% to \textbf{82.1\%}. Besides, comparing with encoding coordinates information with feature representations after the pose encoder, applying the coordinates convolutional encoding operation over the input image pairs directly gives the best performance. Further, we test two different initialization methods for coordinates convolutional layers which are with random initializations or ImageNet-pretrained~\cite{deng2009imagenet} initialization, respectively. The coordinates convolutional encoding layers which are initialized with ImageNet-pretrained weights give slightly improved performance compared to ones with random weights.

\section{Conclusions}
\label{sec:conclusions}
In this work, a novel monocular self-supervised depth estimation framework, called the \textbf{MonoIndoor++}, has been proposed to predict depth map of a single image in indoor environments. The proposed model consists of three modules: (a) a novel \textit{depth factorization module} with a transformer-based scale regression network which is designed to jointly learn a global depth scale factor and a relative depth map from an input image, (b) a novel \textit{residual pose estimation module} which is proposed to estimate accurate relative camera poses for novel view synthesis of self-supervised training that decomposes a global pose into an initial pose and one or a few residual poses, which in turn improves the performance of the depth model, (c) a \textit{coordinates convolutional encoding module} which is utilized to encode coordinates information explicitly to provide additional cues for the residual pose estimation module. Comprehensive evaluation results and ablation studies have been conducted on a wide-variety of indoor datasets, establishing the state-of-the-art performance and demonstrating the effectiveness and universality of our proposed methods.

\bibliographystyle{IEEEtran}
\bibliography{IEEEexample}

\begin{thebibliography}{10}
\providecommand{\url}[1]{#1}
\csname url@samestyle\endcsname
\providecommand{\newblock}{\relax}
\providecommand{\bibinfo}[2]{#2}
\providecommand{\BIBentrySTDinterwordspacing}{\spaceskip=0pt\relax}
\providecommand{\BIBentryALTinterwordstretchfactor}{4}
\providecommand{\BIBentryALTinterwordspacing}{\spaceskip=\fontdimen2\font plus
\BIBentryALTinterwordstretchfactor\fontdimen3\font minus
  \fontdimen4\font\relax}
\providecommand{\BIBforeignlanguage}[2]{{%
\expandafter\ifx\csname l@#1\endcsname\relax
\typeout{** WARNING: IEEEtran.bst: No hyphenation pattern has been}%
\typeout{** loaded for the language `#1'. Using the pattern for}%
\typeout{** the default language instead.}%
\else
\language=\csname l@#1\endcsname
\fi
#2}}
\providecommand{\BIBdecl}{\relax}
\BIBdecl

\bibitem{Eigen_2015_ICCV}
D.~Eigen and R.~Fergus, ``Predicting depth, surface normals and semantic labels
  with a common multi-scale convolutional architecture,'' in \emph{ICCV}, 2015.

\bibitem{Fu_2018_CVPR}
H.~Fu, M.~Gong, C.~Wang, K.~Batmanghelich, and D.~Tao, ``Deep ordinal
  regression network for monocular depth estimation,'' in \emph{CVPR}, 2018.

\bibitem{garg2016unsupervised}
R.~Garg, V.~K. Bg, G.~Carneiro, and I.~Reid, ``Unsupervised cnn for single view
  depth estimation: Geometry to the rescue,'' in \emph{ECCV}, 2016.

\bibitem{zhou2017unsupervised}
T.~Zhou, M.~Brown, N.~Snavely, and D.~G. Lowe, ``Unsupervised learning of depth
  and ego-motion from video,'' in \emph{CVPR}, 2017.

\bibitem{godard2019digging}
C.~Godard, O.~Mac~Aodha, M.~Firman, and G.~J. Brostow, ``Digging into
  self-supervised monocular depth estimation,'' in \emph{ICCV}, 2019.

\bibitem{guo2018learning}
X.~Guo, H.~Li, S.~Yi, J.~Ren, and X.~Wang, ``Learning monocular depth by
  distilling cross-domain stereo networks,'' in \emph{ECCV}, 2018.

\bibitem{Geiger2012CVPR}
A.~Geiger, P.~Lenz, and R.~Urtasun, ``Are we ready for autonomous driving? the
  kitti vision benchmark suite,'' in \emph{CVPR}, 2012.

\bibitem{zhou2019moving}
J.~Zhou, Y.~Wang, K.~Qin, and W.~Zeng, ``Moving indoor: Unsupervised video
  depth learning in challenging environments,'' in \emph{ICCV}, 2019, pp.
  8618--8627.

\bibitem{zhao2020towards}
W.~Zhao, S.~Liu, Y.~Shu, and Y.-J. Liu, ``Towards better generalization: Joint
  depth-pose learning without posenet,'' in \emph{CVPR}, 2020.

\bibitem{bian2021tpami}
J.~Bian, H.~Zhan, N.~Wang, T.-J. Chin, C.~Shen, and I.~Reid, ``Auto-rectify
  network for unsupervised indoor depth estimation,'' \emph{TPAMI}, 2021.

\bibitem{Yin_2019_ICCV}
W.~Yin, Y.~Liu, C.~Shen, and Y.~Yan, ``Enforcing geometric constraints of
  virtual normal for depth prediction,'' in \emph{ICCV}, 2019.

\bibitem{zou2020learning}
Y.~Zou, P.~Ji, , Q.-H. Tran, J.-B. Huang, and M.~Chandraker, ``Learning
  monocular visual odometry via self-supervised long-term modeling,'' in
  \emph{ECCV}, 2020.

\bibitem{mur2017orb}
R.~Mur-Artal and J.~D. Tard{\'o}s, ``Orb-slam2: An open-source slam system for
  monocular, stereo, and rgb-d cameras,'' \emph{TOR}, vol.~33, no.~5, pp.
  1255--1262, 2017.

\bibitem{Silberman:ECCV12}
N.~Silberman, D.~Hoiem, P.~Kohli, and R.~Fergus, ``Indoor segmentation and
  support inference from rgb-d images,'' in \emph{ECCV}, 2012.

\bibitem{schonberger2016structure}
J.~L. Schonberger and J.-M. Frahm, ``Structure-from-motion revisited,'' in
  \emph{CVPR}, 2016.

\bibitem{dosovitskiy2021an}
A.~Dosovitskiy, L.~Beyer, A.~Kolesnikov, D.~Weissenborn, X.~Zhai,
  T.~Unterthiner, M.~Dehghani, M.~Minderer, G.~Heigold, S.~Gelly, J.~Uszkoreit,
  and N.~Houlsby, ``An image is worth 16x16 words: Transformers for image
  recognition at scale,'' in \emph{ICLR}, 2021.

\bibitem{Ji_2021_ICCV}
P.~Ji, R.~Li, B.~Bhanu, and Y.~Xu, ``Monoindoor: Towards good practice of
  self-supervised monocular depth estimation for indoor environments,'' in
  \emph{ICCV}, 2021.

\bibitem{Burri25012016}
M.~Burri, J.~Nikolic, P.~Gohl, T.~Schneider, J.~Rehder, S.~Omari, M.~W.
  Achtelik, and R.~Siegwart, ``The euroc micro aerial vehicle datasets,''
  \emph{IJRR}, vol.~35, no.~10, pp. 1157--1163, 2016.

\bibitem{dai2017scannet}
A.~Dai, A.~X. Chang, M.~Savva, M.~Halber, T.~Funkhouser, and M.~Nie{\ss}ner,
  ``Scannet: Richly-annotated 3d reconstructions of indoor scenes,'' in
  \emph{CVPR}, 2017.

\bibitem{Shotton_2013_CVPR}
J.~Shotton, B.~Glocker, C.~Zach, S.~Izadi, A.~Criminisi, and A.~Fitzgibbon,
  ``Scene coordinate regression forests for camera relocalization in rgb-d
  images,'' in \emph{CVPR}, 2013.

\bibitem{saxena2008make3d}
A.~Saxena, M.~Sun, and A.~Y. Ng, ``Make3d: Learning 3d scene structure from a
  single still image,'' \emph{TPAMI}, vol.~31, no.~5, pp. 824--840, 2008.

\bibitem{schnberger2016pixelwise}
J.~L. Schönberger, E.~Zheng, M.~Pollefeys, and J.-M. Frahm, ``Pixelwise view
  selection for unstructured multi-view stereo,'' in \emph{ECCV}, 2016.

\bibitem{eigen2014depth}
D.~Eigen, C.~Puhrsch, and R.~Fergus, ``Depth map prediction from a single image
  using a multi-scale deep network,'' \emph{arXiv preprint arXiv:1406.2283},
  2014.

\bibitem{laina2016deeper}
I.~Laina, C.~Rupprecht, V.~Belagiannis, F.~Tombari, and N.~Navab, ``Deeper
  depth prediction with fully convolutional residual networks,'' in \emph{3DV},
  2016.

\bibitem{Bhat_2021_CVPR}
S.~F. Bhat, I.~Alhashim, and P.~Wonka, ``Adabins: Depth estimation using
  adaptive bins,'' in \emph{CVPR}, 2021.

\bibitem{Li_2017_ICCV}
J.~Li, R.~Klein, and A.~Yao, ``A two-streamed network for estimating
  fine-scaled depth maps from single rgb images,'' in \emph{ICCV}, 2017.

\bibitem{9316778}
M.~Song, S.~Lim, and W.~Kim, ``Monocular depth estimation using laplacian
  pyramid-based depth residuals,'' \emph{T-CSVT}, vol.~31, no.~11, pp.
  4381--4393, 2021.

\bibitem{8764412}
Y.~Cao, T.~Zhao, K.~Xian, C.~Shen, Z.~Cao, and S.~Xu, ``Monocular depth
  estimation with augmented ordinal depth relationships,'' \emph{T-CSVT},
  vol.~30, no.~8, pp. 2674--2682, 2020.

\bibitem{8010878}
Y.~Cao, Z.~Wu, and C.~Shen, ``Estimating depth from monocular images as
  classification using deep fully convolutional residual networks,''
  \emph{T-CSVT}, vol.~28, no.~11, pp. 3174--3182, 2018.

\bibitem{ummenhofer2017demon}
B.~Ummenhofer, H.~Zhou, J.~Uhrig, N.~Mayer, E.~Ilg, A.~Dosovitskiy, and
  T.~Brox, ``Demon: Depth and motion network for learning monocular stereo,''
  in \emph{CVPR}, 2017, pp. 5038--5047.

\bibitem{teed2018deepv2d}
Z.~Teed and J.~Deng, ``Deepv2d: Video to depth with differentiable structure
  from motion,'' in \emph{ICLR}, 2018.

\bibitem{li2018megadepth}
Z.~Li and N.~Snavely, ``Megadepth: Learning single-view depth prediction from
  internet photos,'' in \emph{CVPR}, 2018.

\bibitem{li2019learning}
Z.~Li, T.~Dekel, F.~Cole, R.~Tucker, N.~Snavely, C.~Liu, and W.~T. Freeman,
  ``Learning the depths of moving people by watching frozen people,'' in
  \emph{CVPR}, 2019.

\bibitem{teed2021droid}
Z.~Teed and J.~Deng, ``{DROID-SLAM: Deep Visual SLAM for Monocular, Stereo, and
  RGB-D Cameras},'' in \emph{NeurIPS}, 2021.

\bibitem{ranftl2019towards}
R.~Ranftl, K.~Lasinger, D.~Hafner, K.~Schindler, and V.~Koltun, ``Towards
  robust monocular depth estimation: Mixing datasets for zero-shot
  cross-dataset transfer,'' \emph{PAMI}, vol.~44, no.~3, pp. 1623--1637, 2022.

\bibitem{Ranftl_2021_ICCV}
R.~Ranftl, A.~Bochkovskiy, and V.~Koltun, ``Vision transformers for dense
  prediction,'' in \emph{ICCV}, 2021.

\bibitem{godard2017unsupervised}
C.~Godard, O.~Mac~Aodha, and G.~J. Brostow, ``Unsupervised monocular depth
  estimation with left-right consistency,'' in \emph{CVPR}, 2017.

\bibitem{bian2019unsupervised}
J.-W. Bian, Z.~Li, N.~Wang, H.~Zhan, C.~Shen, M.-M. Cheng, and I.~Reid,
  ``Unsupervised scale-consistent depth and ego-motion learning from monocular
  video,'' in \emph{NeurIPS}, 2019.

\bibitem{wang2018learning}
C.~Wang, J.~M. Buenaposada, R.~Zhu, and S.~Lucey, ``Learning depth from
  monocular videos using direct methods,'' in \emph{CVPR}, 2018.

\bibitem{yin2018geonet}
Z.~Yin and J.~Shi, ``Geonet: Unsupervised learning of dense depth, optical flow
  and camera pose,'' in \emph{CVPR}, 2018.

\bibitem{zou2018dfnet}
Y.~Zou, Z.~Luo, and J.-B. Huang, ``{DF-Net}: Unsupervised joint learning of
  depth and flow using cross-task consistency,'' in \emph{ECCV}, 2018.

\bibitem{wang2019recurrent}
R.~Wang, S.~M. Pizer, and J.-M. Frahm, ``Recurrent neural network for (un-)
  supervised learning of monocular video visual odometry and depth,'' in
  \emph{CVPR}, 2019.

\bibitem{tiwari2020pseudo}
L.~Tiwari, P.~Ji, Q.-H. Tran, B.~Zhuang, S.~Anand, and M.~Chandraker, ``Pseudo
  rgb-d for self-improving monocular slam and depth prediction,'' in
  \emph{ECCV}, 2020.

\bibitem{watson2021temporal}
J.~Watson, O.~M. Aodha, V.~Prisacariu, G.~Brostow, and M.~Firman, ``{The
  Temporal Opportunist: Self-Supervised Multi-Frame Monocular Depth},'' in
  \emph{CVPR}, 2021.

\bibitem{bian2020unsupervised}
J.-W. Bian, H.~Zhan, N.~Wang, T.-J. Chin, C.~Shen, and I.~Reid, ``Unsupervised
  depth learning in challenging indoor video: Weak rectification to rescue,''
  \emph{arXiv preprint arXiv:2006.02708}, 2020.

\bibitem{Wang_nonlocalCVPR2018}
X.~Wang, R.~Girshick, A.~Gupta, and K.~He, ``Non-local neural networks,'' in
  \emph{CVPR}, 2018.

\bibitem{NIPS2017_3f5ee243}
A.~Vaswani, N.~Shazeer, N.~Parmar, J.~Uszkoreit, L.~Jones, A.~N. Gomez, L.~u.
  Kaiser, and I.~Polosukhin, ``Attention is all you need,'' in \emph{NeurIPS},
  2017.

\bibitem{liu2021Swin}
Z.~Liu, Y.~Lin, Y.~Cao, H.~Hu, Y.~Wei, Z.~Zhang, S.~Lin, and B.~Guo, ``Swin
  transformer: Hierarchical vision transformer using shifted windows,'' in
  \emph{ICCV}, 2021.

\bibitem{liu2021swinv2}
Z.~Liu, H.~Hu, Y.~Lin, Z.~Yao, Z.~Xie, Y.~Wei, J.~Ning, Y.~Cao, Z.~Zhang,
  L.~Dong, F.~Wei, and B.~Guo, ``Swin transformer v2: Scaling up capacity and
  resolution,'' in \emph{CVPR}, 2022.

\bibitem{NIPS2015_14bfa6bb}
S.~Ren, K.~He, R.~Girshick, and J.~Sun, ``Faster r-cnn: Towards real-time
  object detection with region proposal networks,'' in \emph{NeurIPS}, 2015.

\bibitem{NIPS2015_33ceb07b}
M.~Jaderberg, K.~Simonyan, A.~Zisserman, and K.~Kavukcuoglu, ``Spatial
  transformer networks,'' in \emph{NeurIPS}, 2015.

\bibitem{qi2017pointnet}
C.~R. Qi, H.~Su, K.~Mo, and L.~J. Guibas, ``Pointnet: Deep learning on point
  sets for 3d classification and segmentation,'' \emph{CVPR}, 2017.

\bibitem{qi2017pointnetplusplus}
C.~R. Qi, L.~Yi, H.~Su, and L.~J. Guibas, ``Pointnet++: Deep hierarchical
  feature learning on point sets in a metric space,'' in \emph{NeurIPS}, 2017.

\bibitem{mildenhall2020nerf}
B.~Mildenhall, P.~P. Srinivasan, M.~Tancik, J.~T. Barron, R.~Ramamoorthi, and
  R.~Ng, ``Nerf: Representing scenes as neural radiance fields for view
  synthesis,'' in \emph{ECCV}, 2020.

\bibitem{park2021nerfies}
K.~Park, U.~Sinha, J.~T. Barron, S.~Bouaziz, D.~B. Goldman, S.~M. Seitz, and
  R.~Martin-Brualla, ``Nerfies: Deformable neural radiance fields,'' in
  \emph{ICCV}, 2021.

\bibitem{tancik2022blocknerf}
M.~Tancik, V.~Casser, X.~Yan, S.~Pradhan, B.~Mildenhall, P.~Srinivasan, J.~T.
  Barron, and H.~Kretzschmar, ``{Block-NeRF}: Scalable large scene neural view
  synthesis,'' in \emph{CVPR}, 2022.

\bibitem{NEURIPS2018_60106888}
R.~Liu, J.~Lehman, P.~Molino, F.~Petroski~Such, E.~Frank, A.~Sergeev, and
  J.~Yosinski, ``An intriguing failing of convolutional neural networks and the
  coordconv solution,'' in \emph{NeurIPS}, 2018.

\bibitem{Bello_2019_ICCV}
I.~Bello, B.~Zoph, A.~Vaswani, J.~Shlens, and Q.~V. Le, ``Attention augmented
  convolutional networks,'' in \emph{ICCV}, 2019.

\bibitem{chang2018pyramid}
J.-R. Chang and Y.-S. Chen, ``Pyramid stereo matching network,'' in
  \emph{CVPR}, 2018.

\bibitem{NEURIPS2019_9015}
A.~Paszke, S.~Gross, F.~Massa, A.~Lerer, J.~Bradbury, G.~Chanan, T.~Killeen,
  Z.~Lin, N.~Gimelshein, L.~Antiga, A.~Desmaison, A.~Kopf, E.~Yang, Z.~DeVito,
  M.~Raison, A.~Tejani, S.~Chilamkurthy, B.~Steiner, L.~Fang, J.~Bai, and
  S.~Chintala, ``Pytorch: An imperative style, high-performance deep learning
  library,'' in \emph{NeurIPS}, 2019.

\bibitem{kingma2015adam}
D.~P. Kingma and J.~L. Ba, ``Adam: A method for stochastic gradient descent,''
  in \emph{ICLR}, 2015.

\bibitem{Karsch:TPAMI:14}
K.~Karsch, C.~Liu, and S.~B. Kang, ``Depthtransfer: Depth extraction from video
  using non-parametric sampling,'' \emph{TPAMI}, 2014.

\bibitem{Liu_2014_CVPR}
M.~Liu, M.~Salzmann, and X.~He, ``Discrete-continuous depth estimation from a
  single image,'' in \emph{CVPR}, 2014.

\bibitem{Ladicky_2014_CVPR}
L.~Ladicky, J.~Shi, and M.~Pollefeys, ``Pulling things out of perspective,'' in
  \emph{CVPR}, 2014.

\bibitem{Li_2015_CVPR}
B.~Li, C.~Shen, Y.~Dai, A.~van~den Hengel, and M.~He, ``Depth and surface
  normal estimation from monocular images using regression on deep features and
  hierarchical crfs,'' in \emph{CVPR}, 2015.

\bibitem{Roy_2016_CVPR}
A.~Roy and S.~Todorovic, ``Monocular depth estimation using neural regression
  forest,'' in \emph{CVPR}, 2016.

\bibitem{Liu_2015_CVPR}
F.~Liu, C.~Shen, and G.~Lin, ``Deep convolutional neural fields for depth
  estimation from a single image,'' in \emph{CVPR}, 2015.

\bibitem{Wang_2015_CVPR}
P.~Wang, X.~Shen, Z.~Lin, S.~Cohen, B.~Price, and A.~L. Yuille, ``Towards
  unified depth and semantic prediction from a single image,'' in \emph{CVPR},
  2015.

\bibitem{NIPS2016_f3bd5ad5}
A.~Chakrabarti, J.~Shao, and G.~Shakhnarovich, ``Depth from a single image by
  harmonizing overcomplete local network predictions,'' in \emph{NeurIPS},
  2016.

\bibitem{Fang_2020_WACV}
Z.~Fang, X.~Chen, Y.~Chen, and L.~V. Gool, ``Towards good practice for
  cnn-based monocular depth estimation,'' in \emph{WACV}, 2020.

\bibitem{IndoorSfMLearner}
Z.~Yu, L.~Jin, and S.~Gao, ``P$^{2}$net: Patch-match and plane-regularization
  for unsupervised indoor depth estimation,'' in \emph{ECCV}, 2020.

\bibitem{zhan2019dfvo}
H.~Zhan, C.~S. Weerasekera, J.-W. Bian, and I.~Reid, ``Visual odometry
  revisited: What should be learnt?'' in \emph{ICRA}, 2020.

\bibitem{tang2018ba}
C.~Tang and P.~Tan, ``Ba-net: Dense bundle adjustment network,'' 2018.

\bibitem{gu2021dro}
X.~Gu, W.~Yuan, Z.~Dai, C.~Tang, S.~Zhu, and P.~Tan, ``Dro: Deep recurrent
  optimizer for structure-from-motion,'' \emph{arXiv preprint
  arXiv:2103.13201}, 2021.

\bibitem{ummenhofer_demon}
B.~Ummenhofer, H.~Zhou, J.~Uhrig, N.~Mayer, E.~Ilg, A.~Dosovitskiy, and
  T.~Brox, ``Demon: Depth and motion network for learning monocular stereo,''
  in \emph{CVPR}, 2017.

\bibitem{AlismailBL16b}
H.~Alismail, B.~Browning, and S.~Lucey, ``Photometric bundle adjustment for
  vision-based {SLAM},'' \emph{CoRR}, 2016.

\bibitem{sun2021neucon}
J.~Sun, Y.~Xie, L.~Chen, X.~Zhou, and H.~Bao, ``{NeuralRecon}: Real-time
  coherent {3D} reconstruction from monocular video,'' \emph{CVPR}, 2021.

\bibitem{He_2016_CVPR}
K.~He, X.~Zhang, S.~Ren, and J.~Sun, ``Deep residual learning for image
  recognition,'' in \emph{CVPR}, 2016.

\bibitem{deng2009imagenet}
J.~Deng, W.~Dong, R.~Socher, L.-J. Li, K.~Li, and L.~Fei-Fei, ``Imagenet: A
  large-scale hierarchical image database,'' in \emph{CVPR}, 2009.

\end{thebibliography}

\end{document}